\def\year{2022}\relax
\documentclass[letterpaper]{article} 
\usepackage{aaai22}  
\usepackage{times}  
\usepackage{helvet}  
\usepackage{courier}  
\usepackage[hyphens]{url}  
\usepackage{graphicx} 
\usepackage{natbib}  
\usepackage{caption} 
\DeclareCaptionStyle{ruled}{labelfont=normalfont,labelsep=colon,strut=off} 
\usepackage{algorithm}
\usepackage{algorithmic}

\usepackage{color}
\definecolor{Blue}{rgb}{.1,.1,.8}
\definecolor{Red}{rgb}{1,0,0}
\definecolor{Green}{rgb}{.1,.6,.2}

\usepackage{bm}
\usepackage{array}
\usepackage{placeins}
\usepackage{enumitem}
\usepackage{subcaption}
\usepackage{etoolbox}

\usepackage{booktabs}       
\usepackage{nicefrac}       
\usepackage{microtype}      
\usepackage{multirow}
\usepackage{blindtext}
\usepackage{xfrac}
\usepackage{bm}
\usepackage{amsmath}

\usepackage{ifthen}
\newboolean{ack}
\setboolean{ack}{true}

\usepackage{newfloat}
\usepackage{listings}
\floatstyle{ruled}
\newfloat{listing}{tb}{lst}{}
\floatname{listing}{Listing}
\nocopyright
\title{Disentangling Transfer and Interference in Multi-Domain Learning}
\author{
    Yipeng Zhang\textsuperscript{\rm 1},
    Tyler L. Hayes\textsuperscript{\rm 2},
    Christopher Kanan\textsuperscript{\rm 234}
}
\affiliations{
    
    \textsuperscript{\rm 1}University of Rochester, Rochester, NY, USA
    
    \textsuperscript{\rm 2}Rochester Institute of Technology, Rochester, NY, USA
    
    \textsuperscript{\rm 3}Paige, New York, NY, USA
    
    \textsuperscript{\rm 4}Cornell Tech, New York, NY, USA

    yzh232@u.rochester.edu, \{tlh6792, kanan\}@rit.edu
}

\begin{document}

\maketitle

\begin{abstract}
Humans are incredibly good at transferring knowledge from one domain to another, enabling rapid learning of new tasks. Likewise, transfer learning has enabled enormous success in many computer vision problems using pretraining. However, the benefits of transfer in multi-domain learning, where a network learns multiple tasks defined by different datasets, has not been adequately studied. Learning multiple domains could be beneficial, or these domains could interfere with each other given limited network capacity. Understanding how deep neural networks of varied capacity facilitate transfer across inputs from different distributions is a critical step towards open world learning. In this work, we decipher the conditions where interference and knowledge transfer occur in multi-domain learning. We propose new metrics disentangling interference and transfer, set up experimental protocols, and examine the roles of network capacity, task grouping, and dynamic loss weighting in reducing interference and facilitating transfer.

\end{abstract}

\section{Introduction}
\label{sec:intro}

Convolutional Neural Networks (CNNs) have achieved great success in a variety of computer vision tasks, including image classification, object detection, and semantic segmentation~\cite{zamir2018taskonomy}. Although inputs for a particular task can come from various domains, many studies develop models that only solve one task on a single domain. In contrast, humans and animals learn multiple tasks at the same time and utilize task similarities to make better task-level decisions. Inspired by this phenomenon, \textit{multi-task learning (MTL)} seeks to jointly learn a single model for various tasks, typically on the same input domain~\cite{thrun1996learning}. However, in the real world, visual inputs come from several different domains, where an agent must maintain performance on all domains and facilitate transfer among inputs from similar domains. Thus, \textit{multi-domain learning (MDL)} takes the problem a step further and requires models to learn from multiple tasks covering various domains~\cite{bilen2017universal}.
By jointly learning feature representations, MTL and MDL models can achieve superior per-task performance than models trained on a single task in isolation. This is a result of \textit{positive knowledge transfer}~\cite{kendall2018multi}. Facilitating knowledge transfer is a critical step in developing agents that can improve their performance over time on evolving data streams~\cite{mundt2020wholistic,parisi2019continual}.

\begin{figure}
 \centering
      \includegraphics[width=0.65\linewidth]{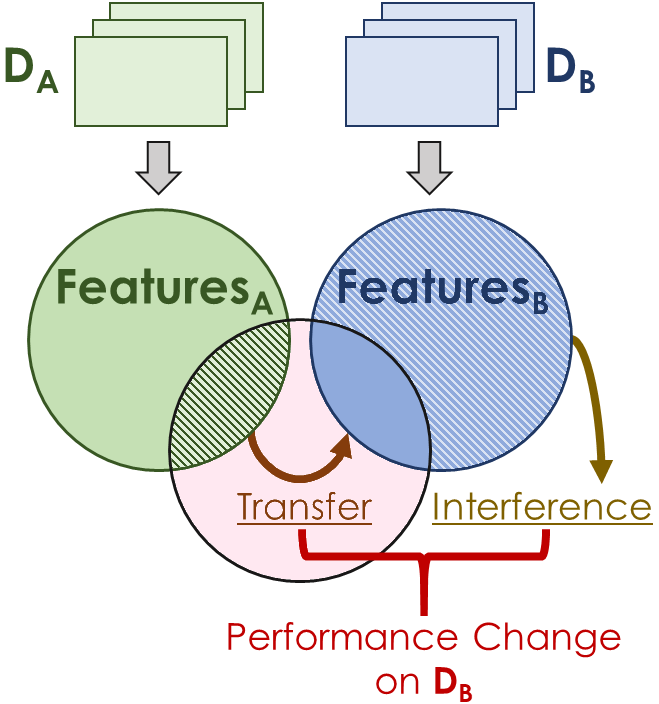}
  \caption{Given two networks trained on domains $D_{A}$ and $D_{B}$ respectively, domain specific features are learned. After jointly training a network on both domains, performance changes on domain $D_B$ are influenced by transfer (shaded green) and interference (shaded blue) from $D_{A}$.
  }
  \label{fig:protocol}
\end{figure}

Unfortunately, jointly training models on multiple tasks does not guarantee performance gains~\cite{zhang2020overcoming, wu2019understanding}. Past work has proposed this can occur due to sample imbalance among tasks~\cite{wu2019understanding, hernandez2021scaling}. We hypothesized that transfer is also affected by a lack of sufficient network capacity to learn all tasks and that task similarity can affect transfer. 

In this paper, we argue that performance differences between an MDL network versus a single-domain independent network stem from the interplay between interference across tasks and the unseen knowledge transferred from other domains/tasks. We propose new metrics and establish experimental protocols to answer this question and empirically evaluate models in the MDL setting using a classification task on three common image datasets. While existing studies have focused on the contribution of task sample sizes in transfer~\cite{wu2019understanding, hernandez2021scaling}, we study three additional factors that could affect the amount of network transfer: 1) network capacity, 2) domain/task groupings, and, 3) task loss weightings~\cite{kendall2018multi, groenendijk2021multi}. In real-world scenarios, making a trade-off between capacity and performance is critical. Intuitively, a model that has more capacity could allocate more task-specific neurons. Further, more similar tasks should yield more transfer and dynamically weighing task-specific losses has been shown to be effective in regulating network learning speed and resolving optimization conflicts of different objectives in MTL~\cite{sermanet2014overfeat, kokkinos2017ubernet, eigen2015predicting, zhang2018context}. Thus, these loss weighting schemes could potentially reduce interference across tasks. Moreover, the effectiveness of loss weighting schemes has not been explored in the MDL setting. 

\textbf{This paper makes the following contributions:}
\begin{enumerate}
\item We propose metrics for task interference and knowledge transfer calculated on a sample-wise basis, which are more comprehensive and robust compared to prior metrics.
\item We introduce an experimental framework for studying knowledge transfer empirically in the multi-domain learning regime, and comprehensively study the individual role of network capacity, task groupings, and multi-domain loss weighting schemes in facilitating knowledge transfer.
\end{enumerate}

\section{Related Work}
\label{sec:related}

\subsection{Multi-Task and Multi-Domain Learning}

Deep MTL architectures can be grouped into hard or soft parameter sharing. Hard parameter sharing models share lower layers of the network, while keeping separate output layers for each task~\cite{thrun1996learning, caruana1997multitask, nekrasov2019real, dvornik2017blitznet, bilen2016integrated, pentina2017multi, doersch2017multi, rudd2016moon}. 
In contrast, soft parameter sharing provides each task with its own network branch~\cite{long2017learning, misra2016cross, dai2016instance, tessler2017deep, lu2017fully, zhou2020pattern}, where information sharing is achieved by adding intermediate modules~\cite{long2017learning,misra2016cross}. 
Some MTL methods focus on weighing different task-specific loss functions, where manually tuning static loss weights has improved performance~\cite{sermanet2014overfeat, kokkinos2017ubernet, eigen2015predicting, zhang2018context}. This is because some tasks may require more attention than others.
However, manually tuning loss weights is not scalable.
Dynamic weighting schemes have become popular due to their scalability and equal effectiveness~\cite{guo2018dynamic, li2017self, kendall2018multi, liebel2018auxiliary, liu2019end, groenendijk2021multi, chen2018gradnorm, sener2018multi}, and can be combined with other architecture modifications~\cite{liu2019end}. We study two weighting methods~\cite{kendall2018multi, groenendijk2021multi}, which we discuss in Sec.~\ref{sec:weighting-schemes}.

Different from MTL, \textbf{multi-domain learning (MDL)} seeks to learn the same types of tasks (e.g., classification), but each task is composed of a different domain~\cite{bilen2017universal}. MDL models seek to achieve high performance on all tasks. Here, we define each task as a separate domain.
During inference, an MDL model is given a data sample with an associated domain label and it performs only the task associated with the indicated domain. Different from MTL, the feature extractor is required to capture different input distributions in the same feature space. Like MTL models, MDL models fall into the hard~\cite{bilen2017universal, rebuffi2017learning, rebuffi2018efficient, berriel2019budget, mallya2018piggyback, bulat2020incremental, mancini2018adding} or soft~\cite{rosenfeld2018incremental, guo2019depthwise, li2019efficient, lee2019learning} parameter sharing categories. However, MDL models must align the feature spaces of multiple domains that contain different lower-level features. Thus, hard sharing is often not performed strictly, but rather by adapting model weights according to domain-specific adaptor modules~\cite{rebuffi2017learning, rebuffi2018efficient} or learning domain-specific weight masks~\cite{mallya2018piggyback, berriel2019budget, mancini2018adding}. Techniques are developed to also enhance feature sharing across domains~\cite{rebuffi2017learning, rebuffi2018efficient, rosenfeld2018incremental}. While most MDL studies seek to design models that facilitate ``transfer,'' defined solely by performance improvement, empirical studies demonstrating the contribution of different factors to performance are missing.

MTL and MDL are closely related to continual learning, which requires models to sequentially learn new tasks~\cite{parisi2019continual,mirzadeh2020linear,van2019three}, instead of jointly learning them. MTL and MDL are also related to transfer learning~\cite{zhuang2020comprehensive}, which aims to improve a model's performance on a target task after pretraining on a source task. Knowledge learned during pretraining can benefit the model in learning the target task. Domain adaptation~\cite{wang2018deep} is a subcategory of transfer learning where the input domain changes and the model must maintain performance on different domains.

\subsection{Knowledge Transfer}

Knowledge transfer is a primary goal in MTL and MDL. An MTL or MDL model exhibits positive transfer when the jointly trained network outperforms a network trained independently on the corresponding task.
Based on this phenomenon, \citet{liebel2018auxiliary} introduced auxiliary tasks, which have been shown to be effective in boosting performance on the main task at a low cost of task construction. 

Knowledge transfer to the target task is the main objective in transfer learning. There have been studies examining the relationship between MTL performance and task transfer in a transfer learning setting~\cite{zamir2018taskonomy, dwivedi2019representation}. However, \citet{standley2020tasks} found that there was no correlation between MTL performance and transfer learning performance. Knowledge transfer has also been studied in continual learning, where a model's ability to perform better on previous tasks (backward transfer) and unseen tasks (forward transfer) is measured~\cite{lopez2017gradient, chaudhry2018riemannian, diaz2018don}. More generally, \citet{standley2020tasks} performed an empirical analysis of transfer in MTL by proposing methods to find the best performing task combinations under a specified computation budget. Here, we go beyond simple measures of network performance to study transfer by introducing metrics to capture interference and transfer separately. Further, the MDL setting allows us to study the role task similarity plays in transfer.

\section{Experimental Protocols}
\label{sec:protocols}

\subsection{Problem Formulation}

We provide a model with a set of classification tasks $\{\mathcal{T}_t\}_{t \leq T}$, along with a set of domains $\{\mathcal{D}_t\}_{t \leq T}$, where $t \leq T$ is a task label. Each task contains a set of $N_{t}$ samples, $\{(\mathbf{x}_i, y_i, t)\}_{i=1}^{N_{t}}$, where $\mathbf{x}_{i}$ is an image, $y_{i}$ is the corresponding label, and $t$ is the task label. 
Our models fall into the hard parameter sharing family of MDL models, where all tasks share the same base model parameters $\bm \theta_{share}$, but contain separate output heads $\bm \theta_{1 \cdots T}$ for tasks $\mathcal{T}_{1 \cdots T}$, respectively. All tasks are trained jointly. We optimize the set of model parameters, ($\bm\theta_{share}^*$, $\bm\theta_{1 \cdots T}^*)$, by minimizing the following loss:
\begin{equation}
    \mathcal{L}_{total} \left( \bm\theta_{share}, \bm\theta_{1 \cdots T} \right) =\sum_{t \leq T} \lambda_t \mathcal{L}_t \left( \bm\theta_{share}, \bm\theta_t\right) \enspace ,
\end{equation}
where $\mathcal{L}_t (\bm\theta_{share}, \bm\theta_t)$ is the loss for task $t$ and $\lambda_t$ is the weight on $\mathcal{L}_t$ which regulates the importance of each task during training. 
Note that for each sample, we provide task labels to the models during calculation of the task-specific losses, so $\mathcal{L}_t$ is only calculated based on outputs from $\bm\theta_t$ without any information from other classification heads.

\subsection{Multi-Domain Loss Weighting Methods}
\label{sec:weighting-schemes}

We investigate the effectiveness of widely-used loss weighting schemes on task losses $\mathcal{L}_{1 \cdots T}$, where each $\mathcal{L}_{t}$ is a standard cross-entropy loss. While these loss weighting schemes have mostly been studied in MTL, we extend them to the MDL setting and describe them below.
    
\textbf{Uniform} -- In Uniform weighting, we use equal task weights, i.e., $\lambda_t$ is set to 1.
    
\textbf{Uncertainty} -- The uncertainty method~\cite{kendall2018multi} models multiple classification objectives with softmax likelihoods by introducing task-dependent uncertainty. Intuitively, this metric captures the relative confidence between tasks, is simple to implement, and is a popular baseline in MTL literature. Task weights $\lambda_t$ are defined as $\sfrac{1}{\varepsilon_t^2}$ where $\varepsilon_{1 \cdots T}$ are learnable noise parameters. Following \cite{liebel2018auxiliary}, we add $\sum_{t \leq T} \log \left(1 + \varepsilon_t^2\right)$ to the loss to avoid trivial solutions.

\textbf{Coefficient of Variations (CoV)} -- CoV is a recent method that achieves state-of-the-art performance in single-task multi-loss settings~\cite{groenendijk2021multi}. Loss weights are estimated based on the variance of single-task loss values in relation to their mean. We use the full history of loss statistics for calculations as described in \cite{groenendijk2021multi}. Because CoV's loss weights sum to one, we multiply them by $T$ to account for scale differences with other methods.

\subsection{Datasets}
\label{sec:datasets}

We perform experiments on three natural image datasets - CIFAR-100 (object/scenery images)~\cite{krizhevsky2009learning}, MiniPlaces (scenery images)~\cite{zhou2017places}, and Tiny-ImageNet (object/scenery images)~\cite{le2015tiny}.
Prior studies have shown that transfer is sensitive to the number of samples in each task~\cite{wu2019understanding, hernandez2021scaling}, so we fix all domains to have an equal number of samples and classes to focus on other factors influencing transfer. Specifically, we use all categories from CIFAR-100 and MiniPlaces and randomly sample a fixed set of 100 categories from Tiny-ImageNet. We then sample a fixed set of 500 training images for each MiniPlaces category.
We resize images from MiniPlaces and Tiny-ImageNet to $32\times32$ pixels using bicubic interpolation. 
We use all test samples pertaining to chosen categories in evaluation. Previous MDL studies~\cite{bilen2017universal, rebuffi2017learning} perform round-robin batch training, where each batch comes from a single dataset. We instead combine all datasets for training, meaning each batch could contain samples from any dataset.

\subsection{Metrics}
\paragraph{Performance:} In prior MDL studies, transfer is quantified as the raw accuracy gain as compared to a single-task baseline~\cite{zamir2018taskonomy, dwivedi2019representation, standley2020tasks}. This metric is insufficient since performance gain consists of both \textit{transfer} and \textit{interference}. In this setting, we cannot infer whether performance gain coincides directly with transfer, so we introduce new metrics.

Suppose we have a model, $\mathcal{M}_{\mathcal{T}_t}$, trained on a single domain, $\mathcal{T}_t$. Given a test set $\mathcal{D}_{test}$,
we denote the set of samples predicted correctly by $\mathcal{M}_{\mathcal{T}_t}$ as $\mathcal{D}_{test}^{correct} \subseteq \mathcal{D}_{test}$ and the samples predicted incorrectly as $\mathcal{D}_{test}^{incorrect} = \mathcal{D}_{test} \setminus \mathcal{D}_{test}^{correct}$. Suppose that an MDL model
makes $k$ and $k'$ correct predictions on $\mathcal{D}_{test}^{correct}$ and $\mathcal{D}_{test}^{incorrect}$, respectively. We define the following metrics to examine its performance on $\mathcal{T}_t$:

\begin{equation}
\textbf{PerfGain} = \frac{k + k' - |\mathcal{D}_{test}^{correct}|}{|\mathcal{D}_{test}|} \times 100\% \enspace ,
\end{equation}
\begin{equation}
\textbf{Interference} = \frac{|\mathcal{D}_{test}^{correct}| - k}{|\mathcal{D}_{test}^{correct}|} \times 100\% \enspace , \end{equation}
\begin{equation}
\textbf{Transfer} = \frac{k'}{|\mathcal{D}_{test}^{incorrect}|} \times 100\% \enspace .
\end{equation}

Performance Gain (PerfGain) is similar to raw performance change metrics used by previous work, but it is measured relative to the single-domain model's performance.
Our transfer score, which is the percentage of samples the MDL model answers correctly among those that the single-domain model fails to solve, indicates how much performance improves by jointly training with other domains. Conversely, interference computes the percentage of samples that the MDL model forgets how to solve among those that the single-domain model answers correctly. It measures the amount of performance degradation due to joint training.

\paragraph{Task Similarity:}

We hypothesize that task similarity in the natural world plays a role in transfer. Since MDL models learn shared feature spaces to simultaneously perform well on many tasks, we calculate task similarity based on feature similarity. That is, given two tasks,
$\mathcal{T}_1$ and $\mathcal{T}_2$, we compute the representational similarity between the associated single-domain models $\mathcal{M}_{\mathcal{T}_1}$ and $\mathcal{M}_{\mathcal{T}_2}$. We evaluate the output of the two models on a fixed dataset and obtain two sets of representations, $X_1$ and $X_2$. This dataset consists of 50 test samples from each class in each dataset used in our experiments. 
We define the similarity between $\mathcal{T}_1$ and $\mathcal{T}_2$ at a particular capacity $w$ as $\text{Sim}_w(\mathcal{T}_1, \mathcal{T}_2)$ using the linear Centered Kernel Alignment (CKA)~\cite{kornblith2019similarity} score between $X_1$ and $X_2$. CKA is a popular metric that has demonstrated robustness.

\subsection{Network Architectures}

We use the ResNet-32 architecture~\cite{he2016deep} for all experiments. We study the effect of network capacity on knowledge transfer by gradually increasing the percentage of neurons used at each layer (i.e., network width), with the exception of prediction heads. We use four widths in our experiments, with $0.25\times$, $0.5\times$, $1\times$, and $2\times$ of the neurons present at each layer. These models contain 29K, 116K, 463K, and 1848K parameters respectively.
In initial studies, ResNet-32 models wider than $2\times$ wide suffered from overfitting, so we do not study them.

Previous work often uses a pretrained model for MDL experiments~\cite{rebuffi2017learning, rebuffi2018efficient, berriel2019budget, mancini2018adding}.
However, this makes it difficult to study transfer in isolation since the model could potentially have informed knowledge about new domains from pretraining.
For this reason, we train models from scratch to study transfer. In addition to jointly trained MDL models, we also train an \textit{independent} (single-domain) model separately on each task as a baseline. We provide implementation details and parameter settings for our experiments in Sec.~\ref{sec:hyperparams}.

\section{Results}
\label{sec:results}

The final accuracies for each independent model are in Sec.~\ref{sec:sdl-perf-sup}. We train one MDL network on each pair of tasks (domain pairings) for all three datasets. 
We repeat this process for three trials (with different random initializations) and perform analysis on the results. 
This leads to \textbf{36 independent models} and $3\text{ (trials)}\times4\text{ (widths)}\times3\text{ (task groupings)}\times3\text{ (loss weightings)}=$ \textbf{108 2-task models}. 
We compute metrics for each MDL model on each domain. 
While studying models trained on two domains enables us to analyze the contribution of task similarity more easily, we include the results of \textbf{36 3-task models} in Sec.~\ref{sec:summary-sup}, Sec.~\ref{sec:trans-inter-sup}, and Sec.~\ref{sec:mdl-perf-sup}.
We organize our results by first outlining high-level questions and claims regarding different factors that could influence interference and transfer and then providing results to address them. 
Error bars in figures denote standard errors.

\begin{figure*}[t]
    \centering
    \begin{subfigure}[t]{0.3\textwidth}
        \includegraphics[width=\linewidth]{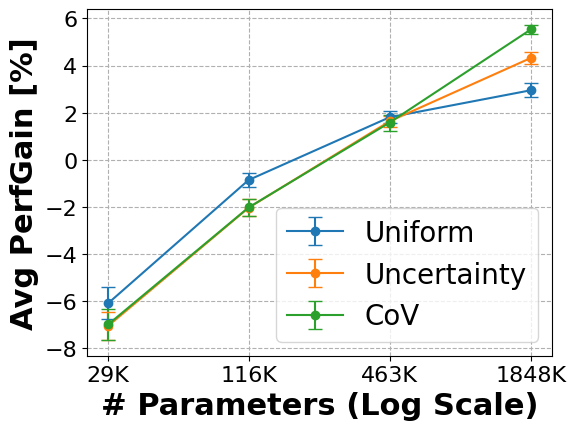}
        \caption{PerfGain}
        \label{fig:summary-perf}
    \end{subfigure} %
    \centering
    \begin{subfigure}[t]{0.3\textwidth}
        \includegraphics[width=\linewidth]{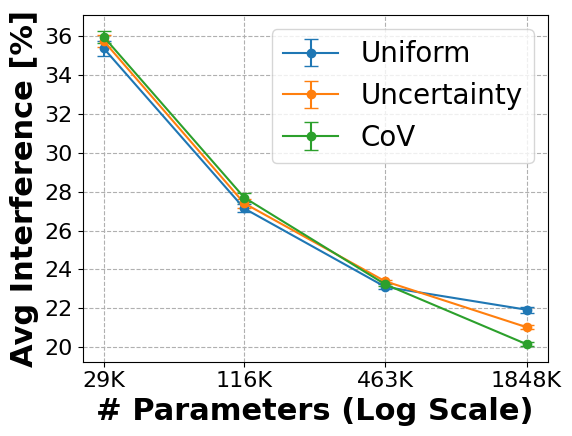}
        \caption{Interference}
        \label{fig:summary-inter}
    \end{subfigure} %
    \centering
    \begin{subfigure}[t]{0.3\textwidth}
        \includegraphics[width=\linewidth]{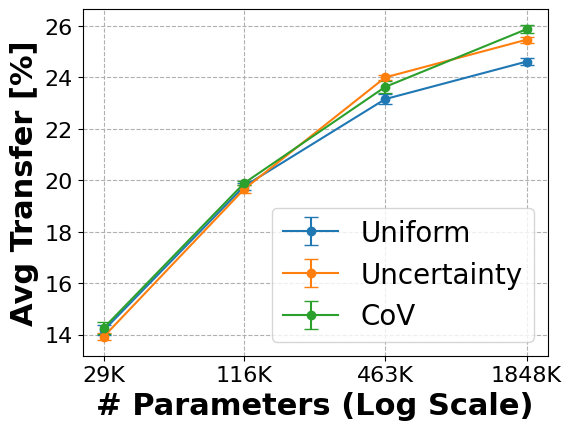}
        \caption{Transfer}
        \label{fig:summary-trans}
    \end{subfigure} %
    \caption{(a) PerfGain, (b) Interference, and (c) Transfer scores averaged over the 2-domain networks at each width, plotted as a function of the network's log number of parameters.}
    \label{fig:summary}
\end{figure*}

\subsection{Correlation Between Transfer Learning and MDL Performance}
\label{sec:tl-mdl}

\begin{table}
\caption{Pearson correlation between transfer learning and MDL models. We underline correlations that are not statistically significant at a 95\% confidence level.
}
\begin{center}
\begin{tabular}{l|ccc}
\toprule
\textbf{Capacity} & \textbf{PerfGain} & \textbf{Transfer} & \textbf{Interference}\\
\midrule
29K & $\underline{0.025}$ & $0.979$ & $0.842$\\
116K & $0.559$ & $0.990$ & $0.982$ \\
463K & $0.505$ & $0.981$ & $0.982$  \\
1848K & $0.739$ & $0.987$ & $0.994$  \\
\bottomrule
\end{tabular}
\end{center}
\label{tab:tl-mdl-corr}
\end{table}

During domain transfer learning, where both the input and label distributions change, a model pretrained on the source domain is optimized towards the target domain alone. The effect of the source domain should remain if we train a model on both domains jointly. \citet{standley2020tasks} found that the performance of standard transfer learning and MTL are not correlated. Curious to see whether the same conclusion holds in MDL, we perform similar experiments more comprehensively with models with different capacities. We first define \textit{directed} task relationship for domains $B \to A$ as the change in performance on $A$ as a consequence of either pretraining (in transfer learning) or jointly training (in MDL) on $B$ and \textit{undirected} task relationship between $A$ and $B$ as the average of both directions.

\textbf{Claim}
\textit{Task relationships in transfer learning positively correlate with that in MDL when characterized by transfer and interference.}

We calculated the directed task relationship on our metrics between all domain pairs at each capacity, under transfer learning and MDL settings and provide undirected results in Sec.~\ref{sec:tl-mdl-corr}. We then check the correlation between the matching relationships in the two settings and include the results in Table~\ref{tab:tl-mdl-corr}. For PerfGain, besides the smallest model, there are significant positive correlations between transfer learning and MDL. This means that the source domain that gives good performance to the target domain in transfer learning is likely to remain beneficial to it if they are trained jointly, especially for large models. This is helpful because transfer learning is less time-consuming than MDL. 

More interestingly, the correlations on transfer or interference are also positive and much stronger, consistent across model sizes. This shows that our sample-wise metrics are more robust and reveal information hidden under the simple accuracy gain. Moreover, such strong correlations allow us to predict the amount of transfer and interference of 2-domain MDL models based on the corresponding transfer learning performance alone.

\subsection{What is the role of dynamic loss weighting?}

Given the success of loss weighting methods for MTL, we want to examine how well they work in the MDL setting, where the input distribution differs for each task. In this section, we focus on overall performance metrics and discuss task similarity metrics in Sec.~\ref{sec:similarity}.

\textbf{Claim}
\textit{With enough capacity, MDL network performance is improved via loss weightings due to a reduction in interference and increase in transfer across tasks.}

Fig.~\ref{fig:summary} shows our metric scores for each network capacity using each loss weighting method, averaged across models. From Fig.~\ref{fig:summary-perf}, we see that dynamic loss weightings only help MDL models for the largest network capacity. This is validated by a paired t-test at each capacity on each pair of the three loss weighting methods. For instance, to perform the test for the Uniform and CoV models, we use PerfGain scores by all Uniform models from all three trials and pair the values with those of the CoV models. For a significance level of $\alpha = 0.05$, we found that the loss weighting methods were only statistically significantly different from one another on models with 1848K parameters, where the CoV method outperforms both the Uniform ($p=0.0165$) and Uncertainty ($p=0.0456$) methods. The dynamic loss weightings are able to reduce interference (Fig.~\ref{fig:summary-inter}) and increase transfer (Fig.~\ref{fig:summary-trans}) at the largest capacity, as verified by a similar paired t-test. For smaller capacities, all three methods have similar interference and transfer scores. We also observe that models using Uniform weighting consistently have a negative PerfGain score on CIFAR-100, while the Uncertainty and CoV methods have positive PerfGain scores when using large network capacities (463K and 1848K parameters). We include scores of individual models in Sec.~\ref{sec:mdl-perf-sup}.

For remaining sections of the paper, we only show results for models using the CoV loss weighting since it performed the best, unless specified otherwise. Results for the Uniform and Uncertainty weightings follow similar trends and are included in Sec.~\ref{sec:mdl-perf-sup}.

\begin{figure*}[t]
    \centering
    \begin{subfigure}[t]{0.3\textwidth}
        \includegraphics[width=\linewidth]{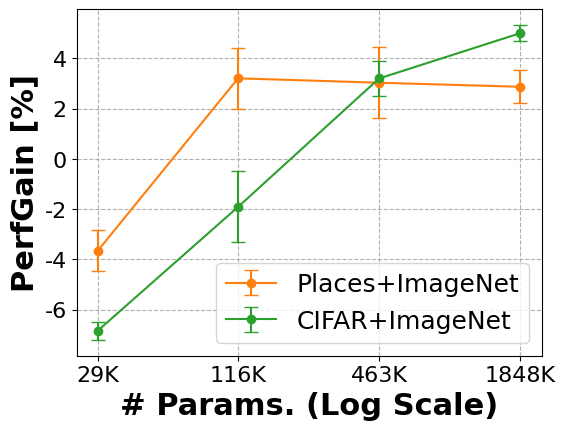}
        \caption{PerfGain}
        \label{fig:metrics-perf}
    \end{subfigure} %
    \centering
    \begin{subfigure}[t]{0.3\textwidth}
        \includegraphics[width=\linewidth]{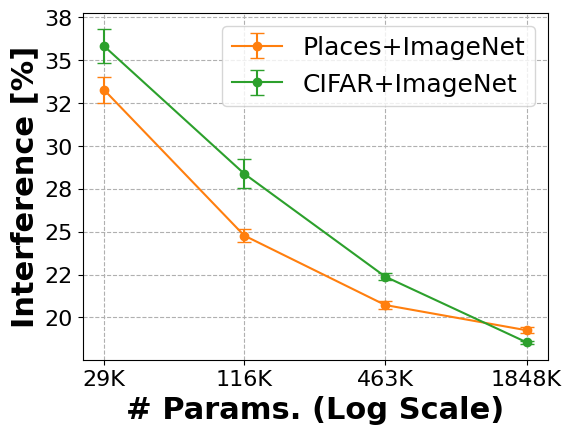}
        \caption{Interference}
        \label{fig:metrics-inter}
    \end{subfigure} %
    \centering
    \begin{subfigure}[t]{0.3\textwidth}
        \includegraphics[width=\linewidth]{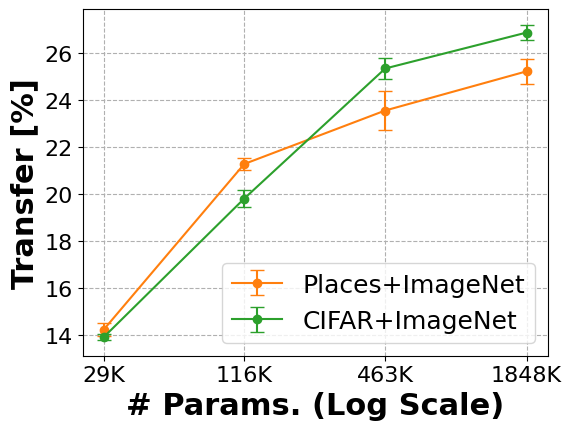}
        \caption{Transfer}
        \label{fig:metrics-trans}
    \end{subfigure} %
    \caption{Our evaluation metrics tested on Tiny-ImageNet.}
    \label{fig:metrics}
\end{figure*}


\subsection{Relationship Between Interference, Transfer, and  Performance}
\label{sec:results-metrics}

MDL models seek to learn a general feature space that is aligned with each of the single-domain model feature spaces. By aligning the general feature space with individual feature spaces, we expect to see the following: (1) the general feature space captures individual task characteristics, which reduces interference, and (2) more information from individual spaces can be adapted to solve other tasks, which increases transfer. 
However, this requires verification because it is possible that a large model uses largely disjoint sets of parameters to learn each task, which would yield little transfer and little interference. We refute this claim by observing linear relationships between transfer and interference (with different slopes in each setting; see Fig.~\ref{fig:inter-trans-sup}). Note that the two scores are calculated on completely disjoint sets of samples.

\textbf{Claim}
\textit{Interference and transfer metrics complement each other.}

We plot the performance gain, interference, and transfer metrics as a function of network capacity when evaluated on Tiny-ImageNet in Fig.~\ref{fig:metrics}.
Consistent across all models, the smallest network exhibits negative PerfGain, but non-zero transfer. Conversely, the largest network has positive PerfGain and non-zero interference. This means that transfer exists when there is performance degradation and interference exists when there is performance improvement. Metrics that only capture overall performance fail to show this. 

For networks with 29K parameters, joint training with MiniPlaces shows larger PerfGain compared to CIFAR-100. We cannot infer that it comes from task interference rather than transfer, unless looking at the other two metrics. Further, the large standard errors of PerfGain prevent us from making definitive conclusions, while our interference and transfer metrics are calculated in a sample-wise fashion and have smaller errors. These observations demonstrate the importance of disentangling the two metrics from overall performance. Further, the discrepancy between these metrics in Sec.~\ref{sec:tl-mdl}, Sec.~\ref{sec:results-capacity}, and Sec.~\ref{sec:similarity} reiterate this claim.


\subsection{What is the role of network capacity?}
\label{sec:results-capacity}

Intuitively, an MDL network with more capacity can allocate more space for storing task-specific information, causing less interference. Conversely, more transfer is expected because larger models can learn how to best share domain-specific information.
However, as mentioned in Sec.~\ref{sec:results-metrics}, it is possible for a large model to not exhibit transfer or interference at all. \citet{wu2019understanding} show that if the capacity of an MTL model is too large then performance might degrade. In MDL, the representations learned from different distributions are expected to be more diverse.

\textbf{Claim}
\textit{Greater capacity leads to less interference and more positive transfer.}

\begin{figure*}[ht]
    \centering
    \begin{subfigure}[t]{0.3\textwidth}
        \includegraphics[width=\linewidth]{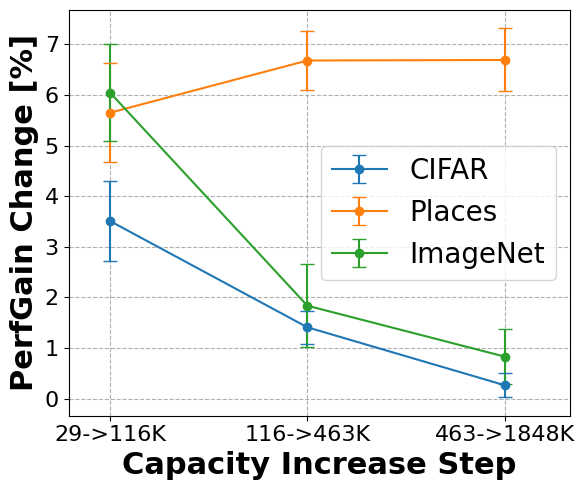}
        \caption{PerfGain}
        \label{fig:pg-change}
    \end{subfigure} %
    \centering
    \begin{subfigure}[t]{0.3\textwidth}
        \includegraphics[width=\linewidth]{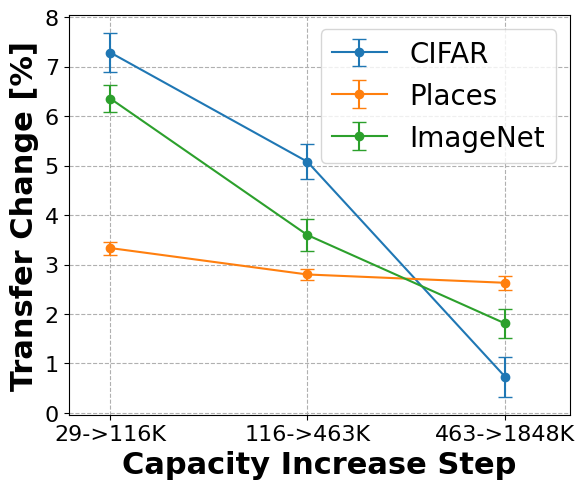}
        \caption{Transfer}
        \label{fig:trans-change}
    \end{subfigure} %
    \centering
    \begin{subfigure}[t]{0.3\textwidth}
        \includegraphics[width=\linewidth]{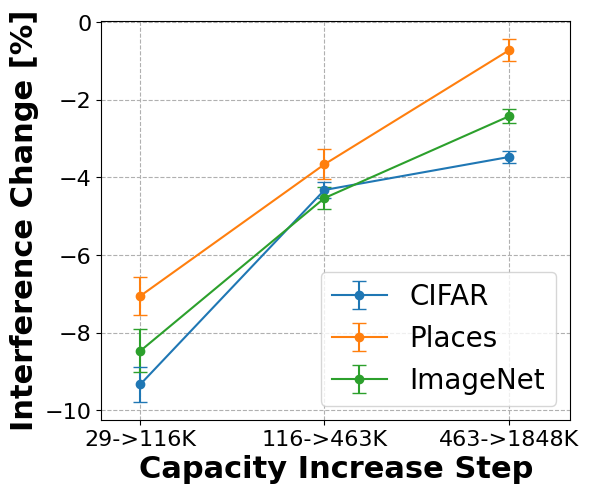}
        \caption{Interference}
        \label{fig:inter-change}
    \end{subfigure} %
    \caption{Average change in (a) PerfGain, (b) Transfer, and (c) Interference for each increment where we increase network capacity. We show the results on each domain, averaged across pairings and loss weighting methods.
    }
    \label{fig:metric-change}
\end{figure*}

Recall that in Fig.~\ref{fig:metrics} we plot our three metrics on Tiny-Imagenet for each pairing. 
We see that the claim is generally true for all models (Fig.~\ref{fig:metrics}, \ref{fig:metrics-naive-sup}, \ref{fig:metrics-unc-sup}, \ref{fig:metrics-cov-sup}).
Our findings in the MDL setting align with those of \citet{wu2019understanding} in very few cases. For example, models having more than 116K parameters trained on MiniPlaces and Tiny-ImageNet experience a slight decrease in PerfGain on Tiny-ImageNet. However, this task pair still has monotonically decreasing interference and increasing transfer. This can be attributed to the fact that resolving interference plays a more important role in overall performance as capacity increases.
Next, we examine transfer on different domains.

\begin{figure*}[t]
    \centering
    \begin{subfigure}[t]{0.3\textwidth}
        \includegraphics[width=\linewidth]{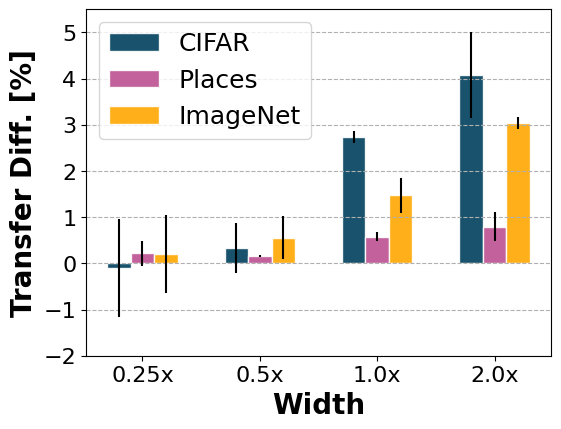}
        \caption{Transfer Diff. (Uniform)}
        \label{fig:transfer-diff-naive}
    \end{subfigure} %
    \centering
    \begin{subfigure}[t]{0.3\textwidth}
        \includegraphics[width=\linewidth]{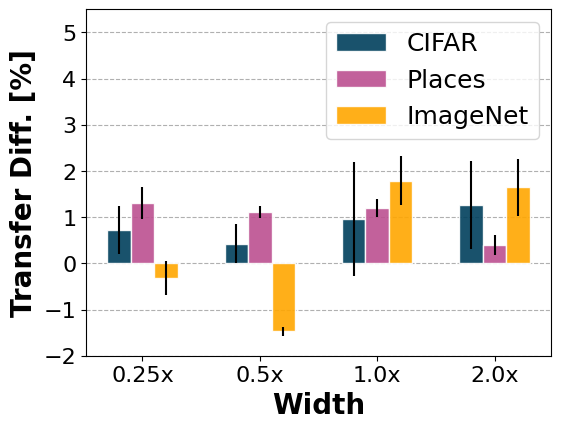}
        \caption{Transfer Diff. (CoV)}
        \label{fig:transfer-diff-cov}
    \end{subfigure} %
    \centering
    \begin{subfigure}[t]{0.3\textwidth}
        \includegraphics[width=\linewidth]{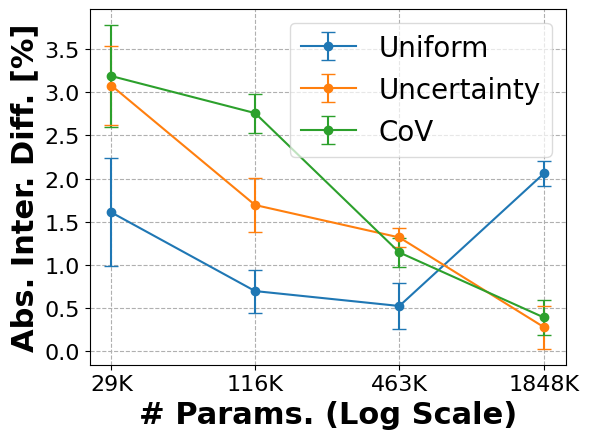}
        \caption{Abs. Interference Diff.}
        \label{fig:avg-inter-diff}
    \end{subfigure} %
    \caption{The difference in transfer when using a more similar task pair than a less similar pair using (a) Uniform and (b) CoV weightings. (c) The absolute difference between interference scores given by the two task pairings, averaged across datasets.
    }
    \label{fig:delta-cka}
\end{figure*}

\textbf{Claim}
\textit{The rate of convergence of transfer on a task is decided by its difficulty.}

Ranked by the single-domain model's accuracy, CIFAR-100 is the easiest task, Tiny-ImageNet is the second easiest, and MiniPlaces is the hardest, consistent across widths. Fig.~\ref{fig:metric-change} plots the change in each of our metrics at each step of capacity increase. Scores are averaged across domain pairings and loss weighting methods.

First, the transfer gain is always positive, but decreases monotonically as capacity grows (\ref{fig:trans-change}). For smaller widths, single-domain networks fail to utilize the space to capture features that are useful to both training and testing distributions, while seeing data from another domain guides the network to learn more general representations. On the other hand, larger networks have space for more features so such guidance for feature selection is less important. Note that transfer gain is always positive because additional domains still contain additional information. Second, the rate of change/convergence of transfer on the three domains (slope of each curve) positively correlates with their single-domain model's accuracy. We cannot make definitive conclusions, but we hypothesize that easier tasks have a higher convergence rate of transfer due to benefit from general features. 

We do not see a clear relationship between the rate of change in PerfGain and task difficulty (Fig.~\ref{fig:pg-change}). The rate of change of interference is independent of task difficulty as well (Fig.~\ref{fig:inter-change}), and is nearly consistent for each domain. The change is monotonically increasing, indicating that the benefit of capacity on interference gradually decreases.

\subsection{How does similarity between tasks affect performance?}
\label{sec:similarity}

For each task pair, we compute our similarity metric and see $\text{Sim}_w(\text{CIFAR}, \text{Places})$ $<$ $\text{Sim}_w(\text{Places}, \text{ImageNet})$ $<$ $\text{Sim}_w(\text{CIFAR}, \text{ImageNet})$ 
consistently for all four widths. 
We now assume this order and show exact scores in Sec.~\ref{sec:sim-analysis-sup}.

In a real-world scenario where we only want to optimize for a particular task
while enforcing less strict requirements for performance on the other task, it is important to know whether we should jointly train with another similar or dissimilar task.
In a similar task pairing, the representations learned by the single-domain models are more similar, meaning it should be easier to create a general feature space when training the two tasks together. Thus, we would expect that more similar task pairs would benefit the original task more.

\textbf{Claim}
\textit{A more similar task pairing yields more transfer, but not necessarily} less interference.

In Fig.~\ref{fig:transfer-diff-naive} and Fig.~\ref{fig:transfer-diff-cov}, we show the amount of transfer gained by using more versus less similar task pairs on each dataset using the Uniform and CoV loss weightings, respectively. A more similar pairing yields more transfer for all the models, violated only by the CoV model's performance on Tiny-ImageNet when using small capacities. Consistent across models with at least 1848K parameters, networks trained with more similar pairings have larger PerfGain scores, at most the same amount of interference, and more transfer (see also Fig.~\ref{fig:metrics}). Although we find that similarity between tasks strongly affects transfer, similarity also reduces interference and improves PerfGain when we use Uniform weighting, but no relationship emerges when we use dynamic weightings (see Sec.~\ref{sec:sim-analysis-sup}). 

\textbf{Claim}
\textit{Dynamic loss weightings are more robust to differing amounts of task similarity.}

Fig.~\ref{fig:transfer-diff-naive} shows that, for Uniform weighting, a more similar pair is more beneficial at larger capacities, while Fig.~\ref{fig:transfer-diff-cov} indicates that there is no clear relationship between similarity and capacity when using dynamic loss weighting. 
We then examine the magnitude of these differences by averaging across their absolute values across the three domains.
Fig.~\ref{fig:avg-inter-diff} examines how the difference between interference changes as capacity increases and shows that networks using loss weighting are less affected by task similarity as size increases.

In summary, dynamic loss weightings yield superior performance at larger capacities (Fig.~\ref{fig:summary}) and, given a domain, reduce the need to determine which other domain to use for joint training. Conversely, if task similarity is known, it is best to choose the most similar domain and use the Uniform weighting model.

\section{Discussion and Conclusion}
\label{sec:discussion}

Our empirical results verify that capacity and similarity heavily influence MDL model performance. \textbf{We summarize our key takeaways as follows.}
\begin{enumerate}
    \item Future studies should perform comparisons of MDL models under various model sizes.
    
    \item Transfer and interference are strongly correlated and, along with performance gain, all saturate exponentially with respect to the growth of capacity.
    
    \item When the network capacity is large enough, dynamic loss weighting methods used in MTL are beneficial to MDL models. These benefits include: more performance gain, less interference, more transfer, and more robustness to task similarity level.
    
    \item With enough capacity, choosing more similar domain pairings guarantees more transfer and PerfGain, and at most, the same amount of interference as a less similar pairing, consistent across datasets and loss weighting methods.
    
    \item Our metrics reveal strong correlations between standard transfer learning and MDL models, allowing us to infer MDL performance from transfer learning performance, and vice versa. Correlation results measured using overall accuracy are not consistent across model capacities.

\end{enumerate}

The non-trivial behaviors of MDL models in exhibiting interference and transfer reiterate the importance of our empirical analysis. 
One future direction our work enables is the study of transfer and interference for additional tasks such as object detection, semantic segmentation, or robotics. While we focused on classification, our metrics are general enough to be applied to other problems. It would be interesting to perform analysis on more challenging datasets such as \cite{rebuffi2017learning}, which involves more dissimilar domains and imbalanced classes. This setting poses further challenges to isolate the impact of each variable on MDL performance.
Moreover, we focused on studying transfer in hard parameter sharing MDL models, but future work could examine the role of transfer in soft parameter sharing methods to determine if our conclusions are consistent across MDL techniques.
Beyond MDL, it would be interesting to apply our metrics in a continual learning setting~\cite{belouadah2020comprehensive,parisi2019continual}.

We established experimental protocols and comprehensive metrics for evaluating task interference and knowledge transfer in MDL. We used these protocols to provide the first empirical study of how capacity, task grouping, and dynamic loss weighting contributes to interference and transfer in the MDL regime. Our results indicate that the interplay between these factors is non-trivial and future studies should not assume how one factor affects another. Studying knowledge transfer in the MDL setting will allow researchers to develop more human-like machines with improved performance. These developments could also help researchers generalize transfer to more challenging settings such as open world learning, where an agent must identify new data and then incrementally learn it~\cite{bendale2015towards,joseph2021towards,xu2019open,mundt2020wholistic}.

\ifthenelse{\boolean{ack}}{
\section*{Acknowledgements}
This work was supported in part by the DARPA/SRI Lifelong Learning Machines program [HR0011-18-C-0051], AFOSR grant [FA9550-18-1-0121], and NSF award \#1909696. The views and conclusions contained herein are those of the authors and should not be interpreted as representing the official policies or endorsements of any sponsor.
}

\bibliography{aaai22}

\appendix

\section{Implementation Details \& Parameter Settings}
\label{sec:hyperparams}

We replace all BatchNorm~\cite{ioffe2015batch} layers in ResNet-32 with GroupNorm~\cite{wu2018group}.
We manually fine-tune the hyper-parameter of GroupNorm (i.e., number of groups) as well as several training hyper-parameters for the highest performance under each setting (i.e., task pairs and widths).
We adopt an adaptive grouping strategy similar to \cite{qiao2019weight} such that each GroupNorm layer has $\min\{32$, $(\text{number of channels})/k\}$ groups. This parameter setting was chosen based on a grid search over $k\in\{2, 4, 8, 16, 32\}$, as well as using a fixed number of groups and searching over $\{1, 4, 8, 16, 32\}$. The adaptive grouping with $k=2$ consistently yielded the best performance over all capacities and task pairs, so we kept it fixed for all experiments.

We also replace all ReLU activation functions with the Mish activation function~\cite{misra2019mish} to prevent gradient vanishing. For all of our experiments, we use a batch size of 128 samples and the stochastic gradient descent (SGD) optimizer with initial learning rate of 0.1, momentum of 0.9, and weight decay of $10^{-4}$. For the single-domain networks, we multiply the learning rate by 0.1 at epochs 140 and 210, and train the model for 250 epochs. For the multi-domain experiments, we multiply the learning rate by 0.1 at epochs 150 and 250, and train the model for 300 epochs.
The learning rate was chosen based on a grid search over the following values: $\{ 0.01, 0.05, 0.1, 0.15, 0.2\}$.

\section{Additional Analysis}

\subsection{Summary Plots for 3-Domain Models}
\label{sec:summary-sup}

\begin{figure*}[ht]
    \centering
    \begin{subfigure}[t]{0.3\textwidth}
        \includegraphics[width=\linewidth]{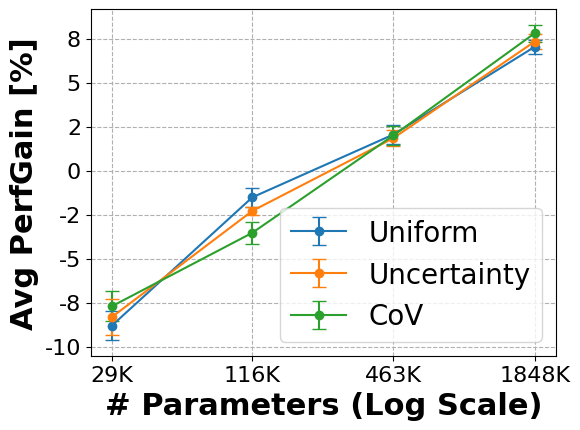}
        \caption{PerfGain}
        \label{fig:summary-perf-3task}
    \end{subfigure} %
    \centering
    \begin{subfigure}[t]{0.3\textwidth}
        \includegraphics[width=\linewidth]{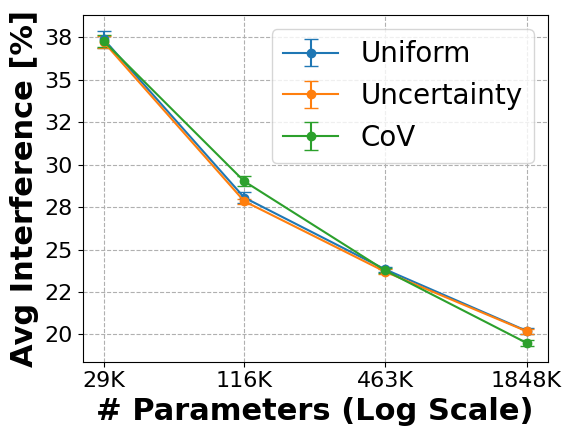}
        \caption{Interference}
        \label{fig:summary-inter-3task}
    \end{subfigure} %
    \centering
    \begin{subfigure}[t]{0.3\textwidth}
        \includegraphics[width=\linewidth]{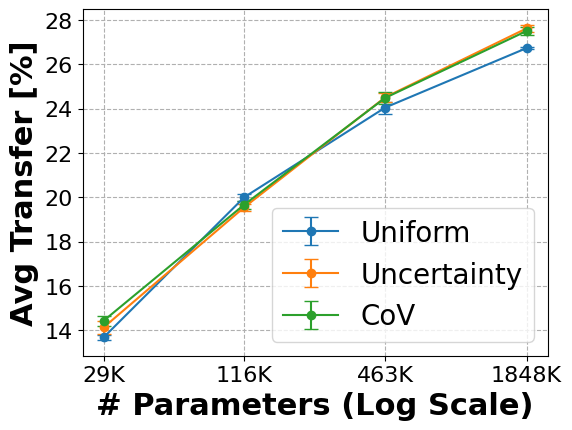}
        \caption{Transfer}
        \label{fig:summary-trans-3task}
    \end{subfigure} %
    \caption{(a) PerfGain, (b) Interference, and (c) Transfer scores averaged over the 3-domain networks at each width, plotted as a function of the network's log number of parameters.}
    \label{fig:summary-3task}
\end{figure*}

Similar to Fig.~\ref{fig:summary}, we plot the average PerfGain, Interference, and Transfer curves as a function of network capacity using 3-domain models (models trained jointly on CIFAR-100, MiniPlaces, and Tiny-ImageNet) in Fig.~\ref{fig:summary-3task}. The general trend in the 2-domain results -- that PerfGain and Transfer increases, and Interference decreases as capacity grows -- holds true for the 3-domain models. However, the 3-domain models' scores saturate slower because training on more domains requires more network capacity to perform well. Unlike the 2-domain PerfGain curve, dynamic loss weighting does not seem to help performance of 3-domain models with 1848K parameters in general.

\subsection{Relationship Between Transfer and Interference}
\label{sec:trans-inter-sup}

\begin{figure*}[ht]
    \centering
    \begin{subfigure}[t]{0.3\textwidth}
        \includegraphics[width=\linewidth]{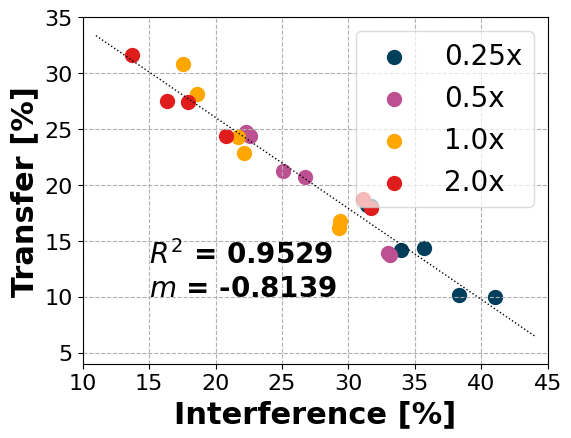}
        \caption{Uniform}
        \label{fig:inter-trans-unif}
    \end{subfigure} %
    \centering
    \begin{subfigure}[t]{0.3\textwidth}
        \includegraphics[width=\linewidth]{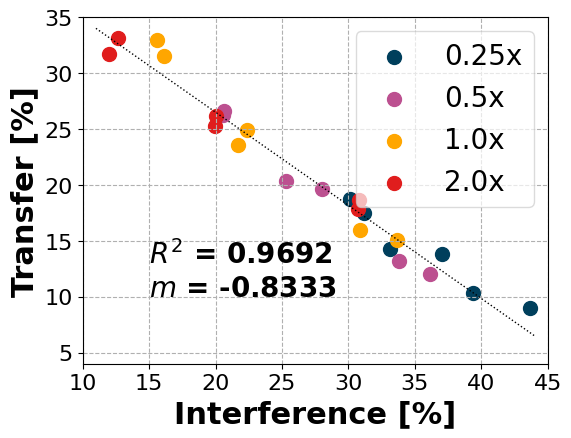}
        \caption{Uncertainty}
        \label{fig:inter-trans-unc}
    \end{subfigure} %
    \centering
    \begin{subfigure}[t]{0.3\textwidth}
        \includegraphics[width=\linewidth]{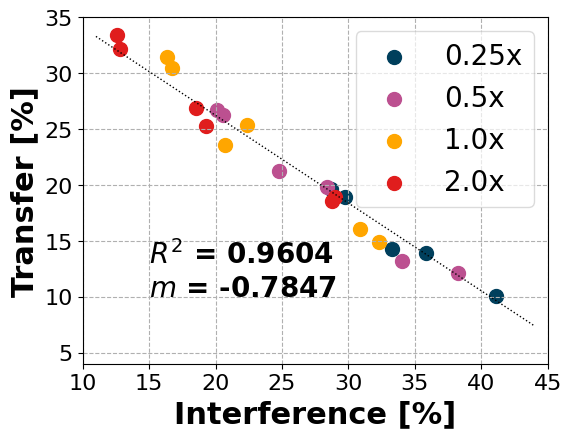}
        \caption{CoV}
        \label{fig:inter-trans-cov}
    \end{subfigure} %
    \caption{Scatter plot of transfer versus interference using (a) Uniform, (b) Uncertainty, and (c) CoV loss weighting with 2-domain models. Each dot corresponds to a network's performance on some domain at some width.  The $R^2$ score is calculated based on the best-fit line, with a slope of $m$. 
    }
    \label{fig:inter-trans-sup}
\end{figure*}

\begin{figure*}[ht]
    \centering
    \begin{subfigure}[t]{0.3\textwidth}
        \includegraphics[width=\linewidth]{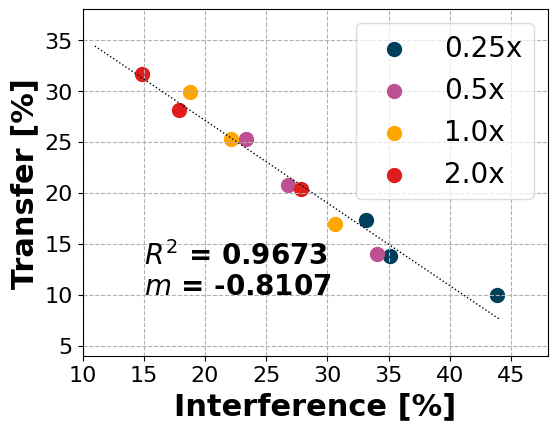}
        \caption{Uniform}
        \label{fig:inter-trans-unif-3task}
    \end{subfigure} %
    \centering
    \begin{subfigure}[t]{0.3\textwidth}
        \includegraphics[width=\linewidth]{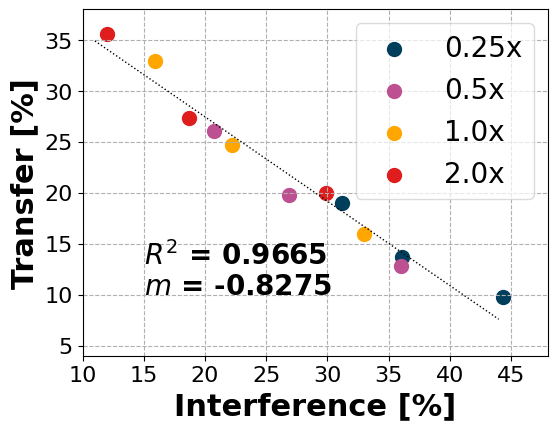}
        \caption{Uncertainty}
        \label{fig:inter-trans-unc-3task}
    \end{subfigure} %
    \centering
    \begin{subfigure}[t]{0.3\textwidth}
        \includegraphics[width=\linewidth]{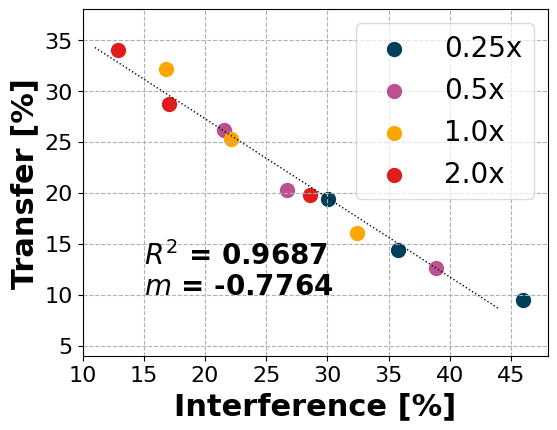}
        \caption{CoV}
        \label{fig:inter-trans-cov-3task}
    \end{subfigure} %
    \caption{Scatter plot of transfer versus interference using (a) Uniform, (b) Uncertainty, and (c) CoV loss weighting with 3-domain models. Each dot corresponds to a network's performance on some domain at some width.  The $R^2$ score is calculated based on the best-fit line, with a slope of $m$.}
    \label{fig:inter-trans-3task-sup}
\end{figure*}

In Fig.~\ref{fig:inter-trans-sup} (2-domain) and Fig.~\ref{fig:inter-trans-3task-sup} (3-domain), we plot each network's transfer as a function of interference using each of the three loss weighting methods. We observe that interference and transfer exhibit a linear relationship, which is consistent across methods. As mentioned in Sec.~\ref{sec:results-metrics}, the correlation strength (slope of the best-fit line) is independent of capacity or domain. Moreover, the correlation strength tells us whether a method focuses more on transfer (more negative slope) or interference (less negative slope). For instance, the best-fit line for CoV has a less negative slope, indicating that it focuses more on transfer.

\subsection{Similarity Analysis}
\label{sec:sim-analysis-sup}

\subsubsection{Task Pairing Similarity Scores}

In Sec.~\ref{sec:similarity}, we use task similarity scores to study interference and transfer. In Table~\ref{tab:cka}, we show the exact CKA similarity scores between each pair of tasks. We observe that the order is consistent across network widths with:
\begin{equation}
\begin{split}
\text{Sim}_w(\text{CIFAR}, \text{Places}) < \\ \text{Sim}_w(\text{Places}, \text{ImageNet}) < \\ \text{Sim}_w(\text{CIFAR}, \text{ImageNet}) \enspace .
\end{split}
\end{equation}

One question that arises is: could the MDL models perform well on each task by only learning task-specific information in the separate classification heads instead of learning generalizable features? Since we have more than one classification head, the shared feature representations would need to be linearly separable in all domains to perform well. This would make it difficult for the model to learn domain-specific features only in the task heads. Further, we verify that the representations learned by the MDL models do not diverge much from those learned by the single domain experts as reflected by CKA scores that were consistently greater than 0.6, which is much higher than the CKA between two single domain experts shown in Table~\ref{tab:cka}. 

\begin{table*}[ht]
\caption{CKA scores of individual models trained on different task pairings at each width of ResNet-32. Scores are listed from the most dissimilar pairings (top) to the most similar pairings (bottom). The mean and standard deviation of CKA scores are reported based on 3 trials with different random initializations.}
\begin{center}
\begin{tabular}{l|cccc}
\toprule
\textbf{Task Pairing}
& $\mathbf{0.25\times}$ & $\mathbf{0.5\times}$ & $\mathbf{1\times}$ & $\mathbf{2\times}$ \\
\midrule
CIFAR \& Places & $0.38 (\pm0.035)$ &  $0.37 (\pm0.147)$ &  $0.42 (\pm0.010)$ &  $0.38 (\pm0.020)$\\
Places \& ImageNet & $0.40(\pm0.027)$ & $0.41(\pm0.009)$ & $0.45(\pm0.010)$ & $0.40(\pm0.023)$ \\
CIFAR \& ImageNet & $0.48 (\pm0.007)$ & $0.50 (\pm0.006)$ & $0.52 (\pm0.008)$ & $0.43(\pm0.002)$ \\
\bottomrule
\end{tabular}
\end{center}
\label{tab:cka}
\end{table*}

\begin{table*}[ht]
\caption{Performance of independent models at each width of ResNet-32. Mean and standard deviation of accuracy scores are shown based on 3 trials with different random initializations.}
\begin{center}
\begin{tabular}{l|cccc}
\toprule
\textbf{Dataset}
& $\mathbf{0.25\times}$ & $\mathbf{0.5\times}$ & $\mathbf{1\times}$ & $\mathbf{2\times}$ \\
\midrule
CIFAR-100 & $42.40 (\pm0.76)$ & $57.12 (\pm0.22)$ & $65.27 (\pm0.23)$ & $70.48 (\pm0.12)$ \\
MiniPlaces & $23.24 (\pm0.58)$ & $29.84 (\pm0.20)$ & $32.04 (\pm0.34)$ & $32.76 (\pm0.34)$ \\
Tiny-ImageNet & $32.46 (\pm0.63)$ & $43.29 (\pm0.74)$ & $49.83 (\pm0.76)$ & $53.28 (\pm0.31)$ \\
\bottomrule
\end{tabular}
\end{center}
\label{tab:ste-perf}
\end{table*}

\subsection{Relationship Between Similarity, Capacity, and Loss Weighting}

\begin{figure*}[t]
    \centering
    \makebox[0.33\textwidth]{\textbf{Uniform}}
    \makebox[0.3\textwidth]{\textbf{Uncertainty}}
    \makebox[0.3\textwidth]{\textbf{CoV}}
    \rotatebox[origin=lt]{90}{\makebox[0.1in]{\textbf{PerfGain Diff.}}}%
    \begin{subfigure}{0.3\textwidth}
        \includegraphics[width=\linewidth]{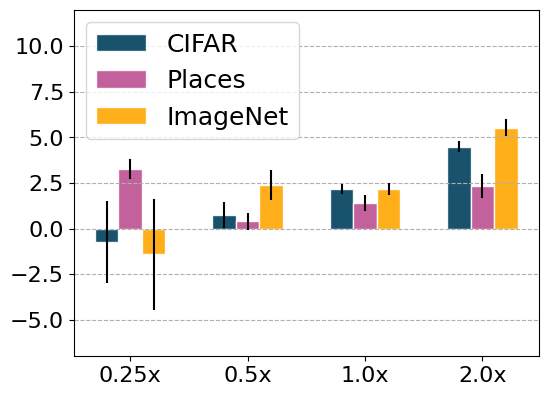}
    \end{subfigure} %
    \centering
    \begin{subfigure}{0.3\textwidth}
        \includegraphics[width=\linewidth]{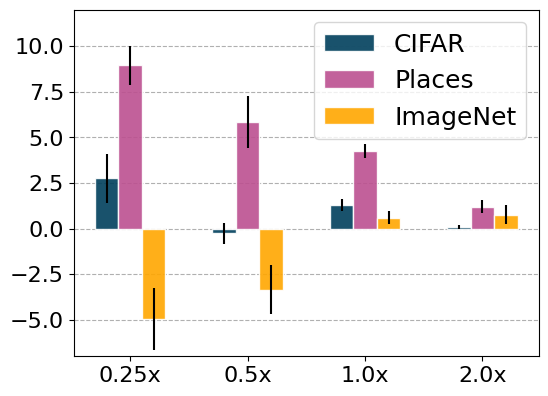}
    \end{subfigure} %
    \centering
    \begin{subfigure}{0.3\textwidth}
        \includegraphics[width=\linewidth]{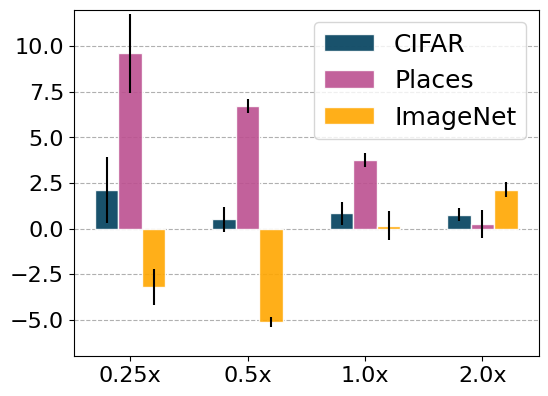}
    \end{subfigure} %

    \rotatebox[origin=lt]{90}{\makebox[0.1in]{\textbf{Transfer Diff.}}}%
    \begin{subfigure}{0.3\textwidth}
        \includegraphics[width=\linewidth]{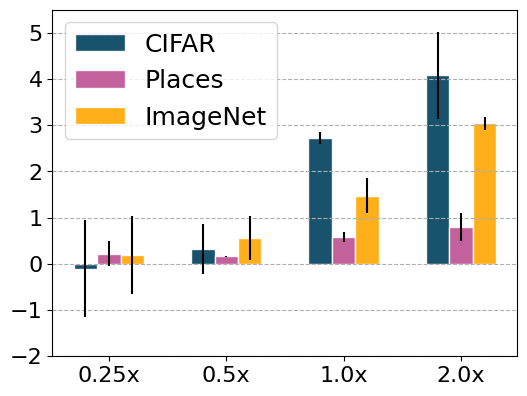}
    \end{subfigure} %
    \centering
    \begin{subfigure}{0.3\textwidth}
        \includegraphics[width=\linewidth]{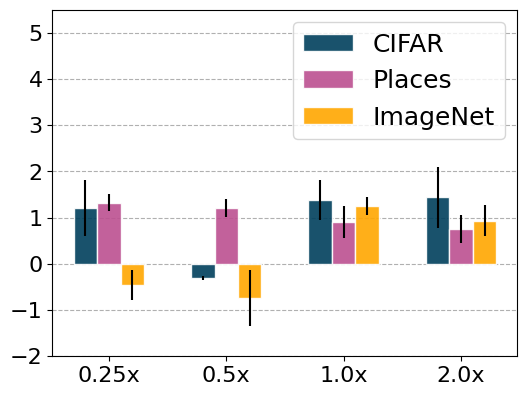}
    \end{subfigure} %
    \centering
    \begin{subfigure}{0.3\textwidth}
        \includegraphics[width=\linewidth]{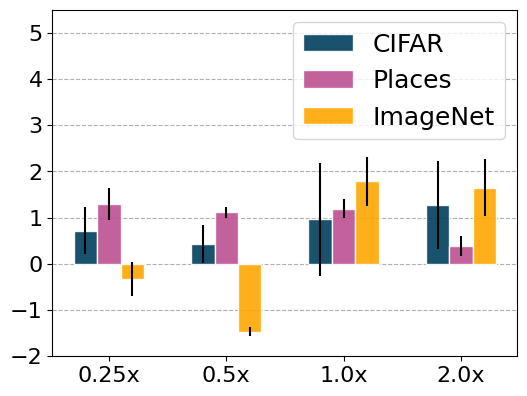}
    \end{subfigure} %

    \rotatebox[origin=lt]{90}{\makebox[0.1in]{\textbf{Interference Diff.}}}%
    \begin{subfigure}{0.3\textwidth}
        \includegraphics[width=\linewidth]{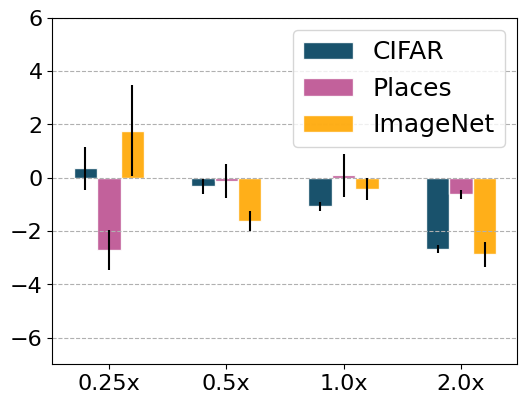}
    \end{subfigure} %
    \centering
    \begin{subfigure}{0.3\textwidth}
        \includegraphics[width=\linewidth]{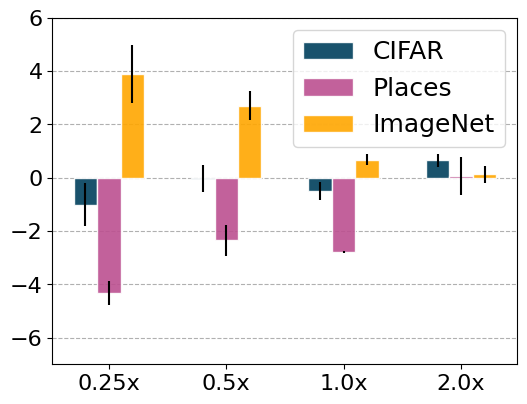}
    \end{subfigure} %
    \centering
    \begin{subfigure}{0.3\textwidth}
        \includegraphics[width=\linewidth]{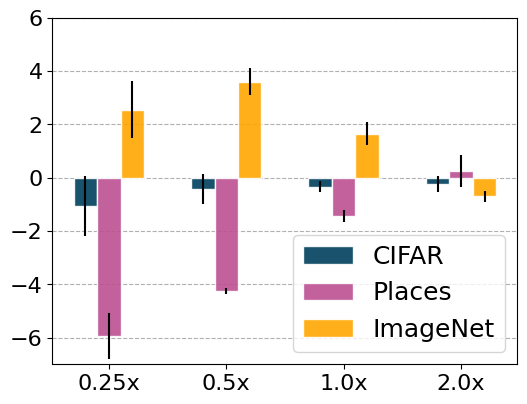}
    \end{subfigure} %

    \makebox[\textwidth]{\textbf{Network Width}}
    
    \caption{The difference in each metric when using a more similar task pair than a less similar pair. The x-axis is the network width with respect to the original ResNet-32.}
    \label{fig:sim-diff-sup}
\end{figure*}

\begin{figure*}[ht]
    \centering
    \begin{subfigure}[t]{0.3\textwidth}
        \includegraphics[width=\linewidth]{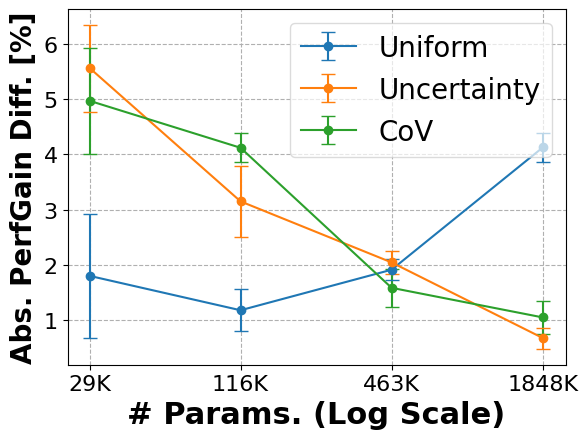}
        \caption{PerfGain}
        \label{fig:abs-pg-diff-sup}
    \end{subfigure} %
    \centering
    \begin{subfigure}[t]{0.3\textwidth}
        \includegraphics[width=\linewidth]{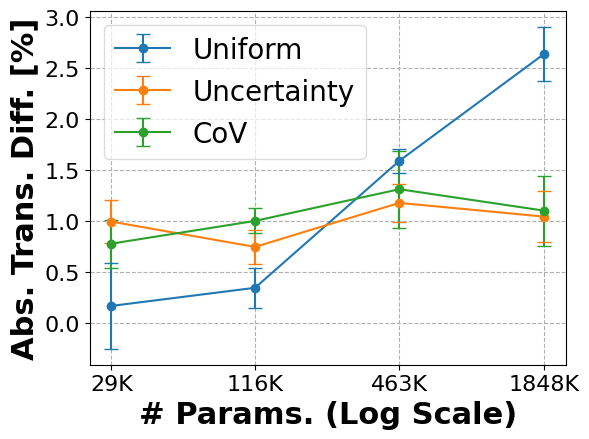}
        \caption{Transfer}
        \label{fig:abs-trans-diff-sup}
    \end{subfigure} %
    \centering
    \begin{subfigure}[t]{0.3\textwidth}
        \includegraphics[width=\linewidth]{images/avg_delta_i_cka.png}
        \caption{Interference}
        \label{fig:abs-inter-diff-sup}
    \end{subfigure} %
    \caption{The absolute difference between (a) PerfGain, (b) Transfer, and (c) Interference given by two task pairs averaged across datasets, plotted against the log number of parameters.
    }
    \label{fig:abs-diff-sup}
\end{figure*}

\subsubsection{Transfer Analysis}

In Fig.~\ref{fig:sim-diff-sup} (middle row), we plot the transfer gain when using a more similar task pairing for 2-domain models. As discussed in the main text, using a more similar task pair with the Uniform method always yields more transfer. For both the Uncertainty and CoV weightings, the conclusion is true except for Tiny-ImageNet at width $0.25\times$ and $0.5\times$, and for CIFAR-100 at width $0.5\times$.
These special cases indicate that networks using loss weightings sometimes fail to utilize the similarity between tasks to facilitate transfer, which is likely due to a lack of capacity.

We then examine the magnitude (absolute value) of this difference, averaged across datasets, as shown in Fig.~\ref{fig:abs-trans-diff-sup}. Using the Uniform weighting, the difference increases as width increases, indicating that it is important to choose the more similar pair when using large capacities to obtain the most transfer. Conversely, using the dynamic loss weighting methods, the absolute difference in transfer between task pairs remains the same regardless of width.

\subsubsection{Interference Analysis}

In Fig.~\ref{fig:sim-diff-sup} (last row), we plot the interference change when using a more similar task pair for 2-domain models. Interestingly, unlike transfer, we do not see a clear trend when considering if the value of the interference difference is positive or negative.
For the Uniform method, the difference is generally negative, indicating that using a more similar task pair generally yields less interference. We do not see a clear relationship between similarity and interference for dynamic loss weightings, where MiniPlaces benefits most from a more similar task pair, while Tiny-ImageNet benefits more from a less similar one. Therefore, jointly training a more similar domain does not necessarily guarantee that there will be less interference than training using a less similar domain. However, we next discuss that similarity does not have to be considered when using loss weighting methods with large capacity networks.

Using the Uniform loss weighting, there is no clear relationship between the magnitude of the interference differences between task pairs and network capacity (Fig.~\ref{fig:abs-inter-diff-sup}). Using dynamic weightings, the magnitude of the interference differences decreases as capacity grows, as shown in the main text (Fig.~\ref{fig:avg-inter-diff}) as well as in Fig.~\ref{fig:abs-inter-diff-sup}. Therefore, although we find no correlation between similarity and interference for dynamic weightings, we show that the differences are negligible when we have a large network.

\subsubsection{PerfGain Analysis}

Performance improvement is a result of the interplay between transfer and interference, and we plot the PerfGain change when using a more similar task pair for 2-domain models (first row of Fig.~\ref{fig:sim-diff-sup}). For the Uniform model, the behavior of PerfGain is consistent with transfer, while for the dynamic weighting models, its behavior is consistent with the negation of interference. 

In terms of the magnitude of PerfGain differences, dynamic loss weighting methods are less affected by the difference in task similarity as capacity grows (Fig.~\ref{fig:abs-pg-diff-sup}). On the other hand, the benefit of a more similar task pair grows as capacity grows if we use the Uniform method.

Similar to our discussion in Sec.~\ref{sec:similarity}, the key takeaway of Fig.~\ref{fig:sim-diff-sup} and Fig.~\ref{fig:abs-diff-sup} is as follows: dynamic loss weightings yield superior performance at larger capacities (Fig.~\ref{fig:summary}) and, given a domain, reduce the need to determine which other domain to use for joint training. On the other hand, if the order of similarity is known, it is best to choose the most similar domain and use the Uniform weighting model.

\section{Transfer Learning and MDL Performance Correlation}
\label{sec:tl-mdl-corr}

\begin{table*}[ht]
\caption{Directed domain relationship correlations between Transfer Learning and Multi-Domain Learning models. We report the Pearson's correlation coefficient and the $p$-value for statistical significance for each experiment.}
\begin{center}
\begin{tabular}{l|cc|cc|cc}
\toprule
\textbf{Capacity} & \multicolumn{2}{c|}{\textbf{PerfGain}} & \multicolumn{2}{c|}{\textbf{Transfer}} & \multicolumn{2}{c}{\textbf{Interference}} \\
\midrule
(\# param.) & $\text{Pearson's } r$ & $p$ & $\text{Pearson's } r$ & $p$ & $\text{Pearson's } r$ & $p$ \\
\midrule
29K & $0.025$ & $0.311$ & $0.979$ & $1\times10^{-12}$ & $0.842$ & $1\times10^{-5}$  \\
116K & $0.559$ & $0.016$ & $0.990$ & $5\times10^{-15}$ & $0.982$ & $5\times10^{-13}$  \\
463K & $0.505$ & $0.033$ & $0.981$ & $6\times10^{-13}$ & $0.982$ & $4\times10^{-13}$  \\
1848K & $0.739$ & $0.001$ & $0.987$ & $4\times10^{-14}$ & $0.994$ & $1\times10^{-16}$  \\
\bottomrule
\end{tabular}
\end{center}
\label{tab:aff-rel-directed}
\end{table*}

\begin{table*}[ht]
\caption{Undirected domain relationship correlations between Transfer Learning and Multi-Domain Learning models. We report the Pearson's correlation coefficient and the $p$-value for statistical significance for each experiment.}
\begin{center}
\begin{tabular}{l|cc|cc|cc}
\toprule
\textbf{Capacity} & \multicolumn{2}{c|}{\textbf{PerfGain}} & \multicolumn{2}{c|}{\textbf{Transfer}} & \multicolumn{2}{c}{\textbf{Interference}} \\
\midrule
(\# param.) & $\text{Pearson's } r$ & $p$ & $\text{Pearson's } r$ & $p$ & $\text{Pearson's } r$ & $p$ \\
\midrule
29K & $0.118$ & $0.762$ & $0.942$ & $1\times10^{-4}$ & $0.573$ & $0.107$  \\
116K & $0.651$ & $0.057$ & $0.986$ & $1\times10^{-6}$ & $0.922$ & $4\times10^{-4}$  \\
463K & $0.142$ & $0.716$ & $0.962$ & $3\times10^{-5}$ & $0.958$ & $4\times10^{-5}$  \\
1848K & $0.222$ & $0.565$ & $0.986$ & $1\times10^{-6}$ & $0.987$ & $8\times10^{-7}$  \\
\bottomrule
\end{tabular}
\end{center}
\label{tab:aff-rel-undirected}
\end{table*}

We include the full table of directed (Table~\ref{tab:aff-rel-directed}) and undirected (Table~\ref{tab:aff-rel-undirected}) domain relationship correlations between Transfer Learning and MDL, including the $p$-values for statistical significance testing. As discussed in Sec.~\ref{sec:tl-mdl}, the directed correlation between Transfer Learning and MDL is positive except for PerfGain using the network with smallest capacity. We also see much stronger and statistically significant correlations between the Transfer Learning and MDL relationships using our transfer and interference metrics. Conversely, we do not see correlation between the Transfer Learning and MDL relationship using our PerfGain scores. However, the all correlations using our transfer and interference metrics are statistically significant at a 95\% confidence interval, with the exception of our interference score for the network with the smallest capacity. The PerfGain correlations are consistent with the findings of \citet{standley2020tasks}, in which experiments are performed on Transfer Learning and MTL with only a single network (i.e., not varying capacity). As shown in Table~\ref{tab:aff-rel-directed} and Table~\ref{tab:aff-rel-undirected}, correlation results vary across capacity, and the results of experiments that were conducted using only a single capacity may not hold for different network capacities or architectures. Our transfer and interference metrics are more robust for revealing the underlying correlations between Transfer Learning and MDL.

\section{Single-Domain Model Performance}
\label{sec:sdl-perf-sup}

In Table~\ref{tab:ste-perf}, we show the final accuracy of each single-domain model trained and evaluated on each domain. As mentioned in Sec.~\ref{sec:results-capacity}, we define the difficulty of tasks based on their respective single-domain model performance. That is, CIFAR-100 is the easiest and Tiny-ImageNet is the hardest, which is consistent across widths. One observation is that performance gains gradually decrease when adding more network capacity on all three datasets.

\section{Multi-Domain Model Performances}
\label{sec:mdl-perf-sup}

We compute our evaluation metrics with networks using Uniform, Uncertainty, and CoV loss weighting, as shown in Fig.~\ref{fig:metrics-naive-sup}, Fig.~\ref{fig:metrics-unc-sup}, and Fig.~\ref{fig:metrics-cov-sup}, respectively. We first discuss the results of 2-domain models.

\subsection{2-Domain Model Performance}

Recall from Sec.~\ref{sec:results}, there are some general trends that are consistent across domains and methods. First, for small capacities, all networks exhibit negative PerfGain scores, while having positive transfer; at the largest capacity, only the Uniform methods show negative PerfGain on CIFAR-100 and all other networks show positive PerfGain, while having positive interference. Second, at the largest capacity (1848K parameters), all models using a more similar task pairing show larger PerfGain, at most the same amount of interference, and more transfer.

The two dynamic loss weighting methods show very similar behaviors. As mentioned in the main text (Fig.~\ref{fig:avg-inter-diff}), the interference difference between task pairs decreases as capacity grows when using dynamic loss weighting. 

Next, we list cases which indicate that a single overall performance gain metric is not enough to capture both transfer and interference. \textbf{(1)} As discussed in Sec.~\ref{sec:results-metrics}, using the $0.25\times$ wide CoV networks, we cannot directly infer that a large PerfGain score on Tiny-ImageNet from jointly training with MiniPlaces is attributed to the task pair's ability to reduce interference, rather than increasing transfer (see right column of Fig.~\ref{fig:metrics-cov-sup}). \textbf{(2)} Similarly, using the $0.25\times$ wide Uniform networks, we cannot directly infer that a large PerfGain on MiniPlaces from jointly training with Tiny-ImageNet is attributed to the task pair's ability to reduce interference, rather than increasing transfer (see middle column of Fig.~\ref{fig:metrics-naive-sup}). \textbf{(3)} Using the Uniform network jointly trained with MiniPlaces (see left column of Fig.~\ref{fig:metrics-naive-sup}), we notice that the PerfGain score on CIFAR-100 decreases at the largest capacity (1848K parameters). Separating the interference and transfer metrics enables us to discover that the network fails to promote more transfer, but still manages to reduce more interference compared to the smaller network at 463K parameters. There are more similar examples in Fig.~\ref{fig:metrics-naive-sup}, Fig.~\ref{fig:metrics-unc-sup}, and Fig.~\ref{fig:metrics-cov-sup}. Our metrics present us with a more comprehensive view of an MDL model's performance, as compared to only evaluating overall performance.

\subsection{3-Domain Model Performance}

\begin{figure*}[t]
    \centering
    \makebox[0.33\textwidth]{\textbf{CIFAR-100}}
    \makebox[0.3\textwidth]{\textbf{MiniPlaces}}
    \makebox[0.3\textwidth]{\textbf{Tiny-ImageNet}}
    \rotatebox[origin=lt]{90}{\makebox[0.1in]{\textbf{PerfGain [\%]}}}%
    \begin{subfigure}{0.3\textwidth}
        \includegraphics[width=\linewidth]{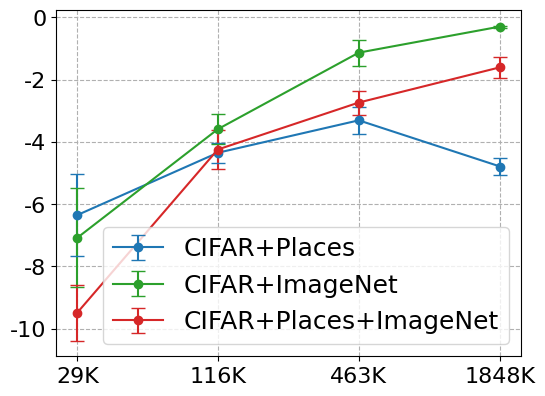}
    \end{subfigure} %
    \centering
    \begin{subfigure}{0.3\textwidth}
        \includegraphics[width=\linewidth]{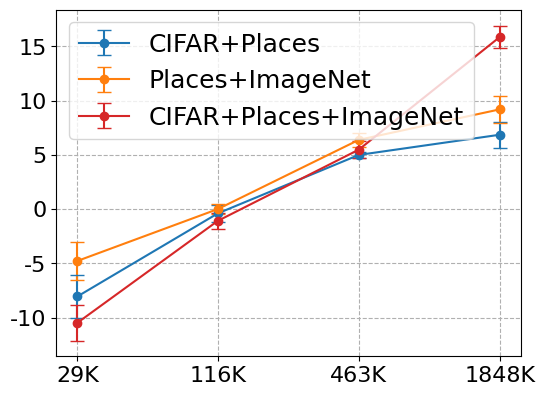}
    \end{subfigure} %
    \centering
    \begin{subfigure}{0.3\textwidth}
        \includegraphics[width=\linewidth]{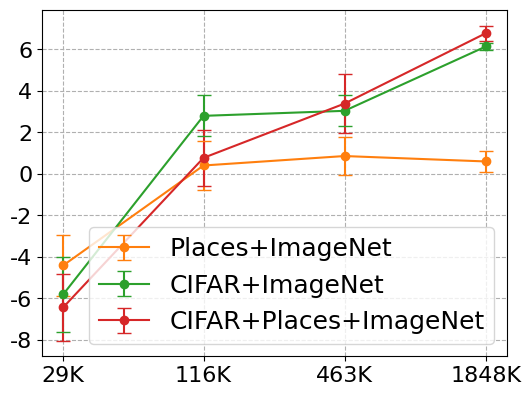}
    \end{subfigure} %

    \rotatebox[origin=lt]{90}{\makebox[0.1in]{\textbf{Interference [\%]}}}%
    \begin{subfigure}{0.3\textwidth}
        \includegraphics[width=\linewidth]{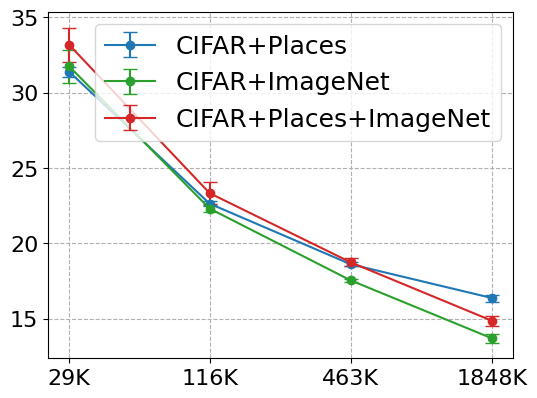}
    \end{subfigure} %
    \centering
    \begin{subfigure}{0.3\textwidth}
        \includegraphics[width=\linewidth]{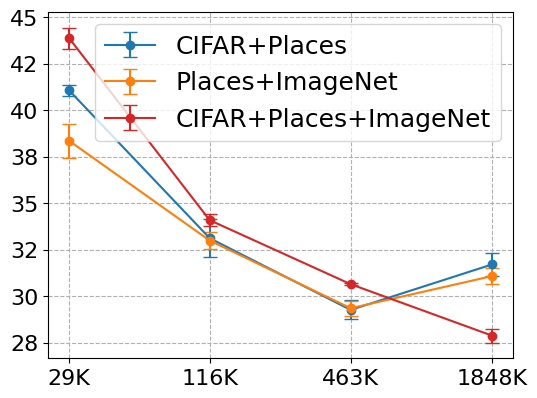}
    \end{subfigure} %
    \centering
    \begin{subfigure}{0.3\textwidth}
        \includegraphics[width=\linewidth]{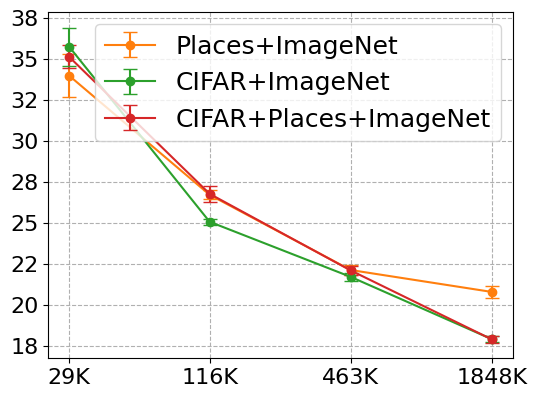}
    \end{subfigure} %

    \rotatebox[origin=lt]{90}{\makebox[0.1in]{\textbf{Transfer [\%]}}}%
    \begin{subfigure}{0.3\textwidth}
        \includegraphics[width=\linewidth]{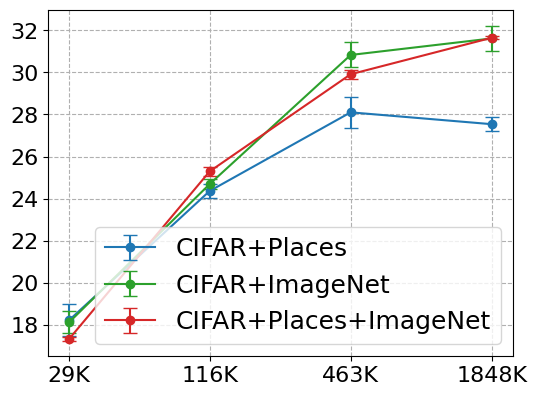}
    \end{subfigure} %
    \centering
    \begin{subfigure}{0.3\textwidth}
        \includegraphics[width=\linewidth]{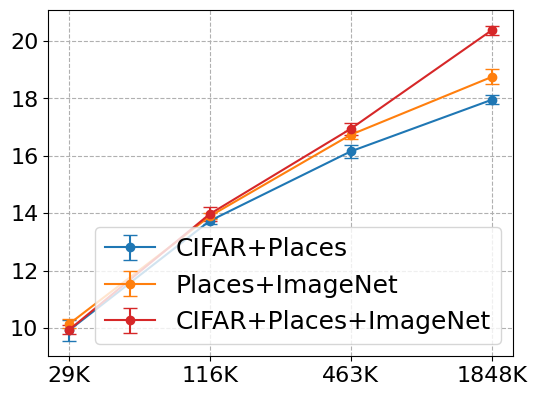}
    \end{subfigure} %
    \centering
    \begin{subfigure}{0.3\textwidth}
        \includegraphics[width=\linewidth]{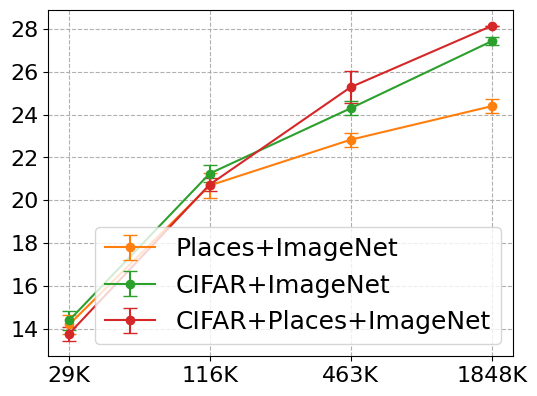}
    \end{subfigure} %

    \makebox[\textwidth]{\textbf{Number of Parameters (Log Scale)}}
    
    \caption{Our evaluation metrics tested with networks using Uniform loss weighting.}
    \label{fig:metrics-naive-sup}
\end{figure*}

\begin{figure*}[t]
    \centering
    \makebox[0.33\textwidth]{\textbf{CIFAR-100}}
    \makebox[0.3\textwidth]{\textbf{MiniPlaces}}
    \makebox[0.3\textwidth]{\textbf{Tiny-ImageNet}}
    \rotatebox[origin=lt]{90}{\makebox[0.1in]{\textbf{PerfGain [\%]}}}%
    \begin{subfigure}{0.3\textwidth}
        \includegraphics[width=\linewidth]{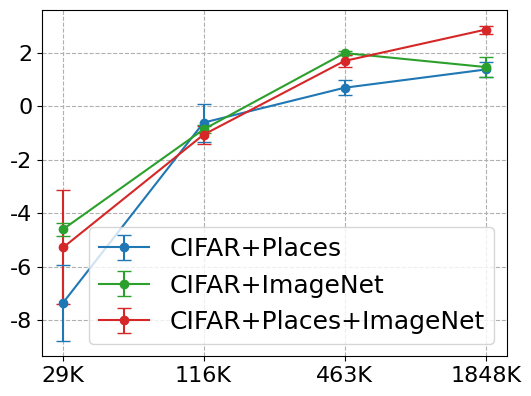}
    \end{subfigure} %
    \centering
    \begin{subfigure}{0.3\textwidth}
        \includegraphics[width=\linewidth]{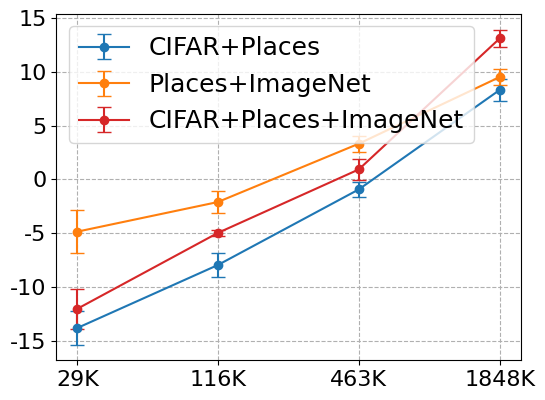}
    \end{subfigure} %
    \centering
    \begin{subfigure}{0.3\textwidth}
        \includegraphics[width=\linewidth]{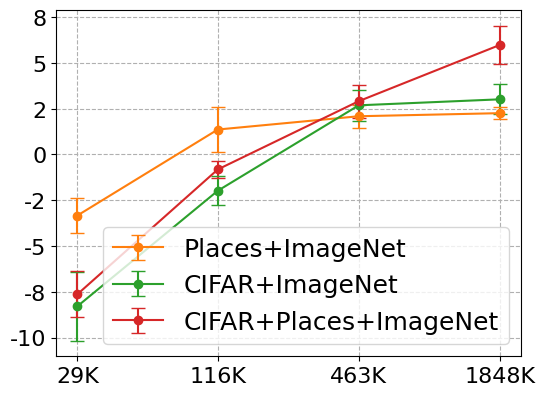}
    \end{subfigure} %

    \rotatebox[origin=lt]{90}{\makebox[0.1in]{\textbf{Interference [\%]}}}%
    \begin{subfigure}{0.3\textwidth}
        \includegraphics[width=\linewidth]{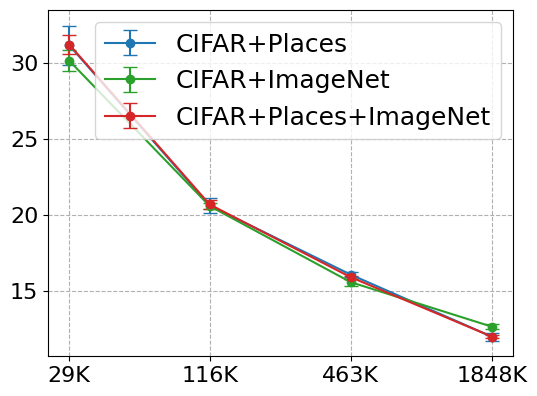}
    \end{subfigure} %
    \centering
    \begin{subfigure}{0.3\textwidth}
        \includegraphics[width=\linewidth]{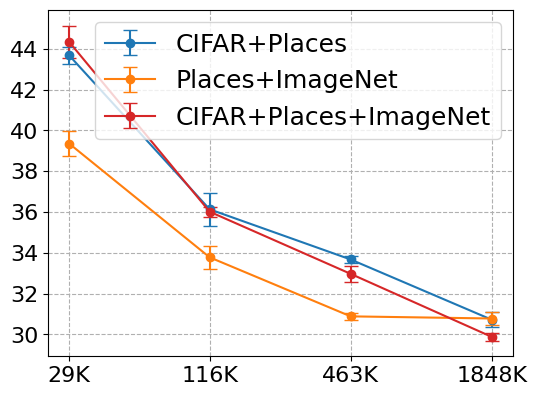}
    \end{subfigure} %
    \centering
    \begin{subfigure}{0.3\textwidth}
        \includegraphics[width=\linewidth]{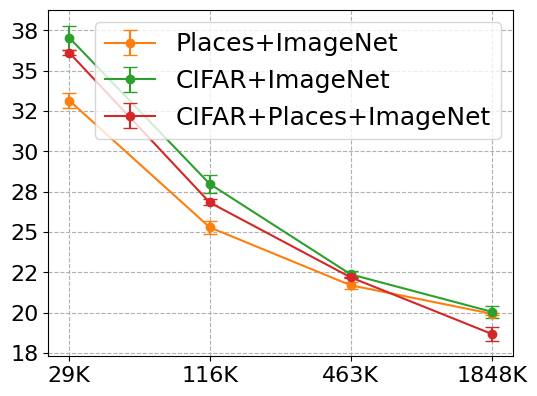}
    \end{subfigure} %

    \rotatebox[origin=lt]{90}{\makebox[0.1in]{\textbf{Transfer [\%]}}}%
    \begin{subfigure}{0.3\textwidth}
        \includegraphics[width=\linewidth]{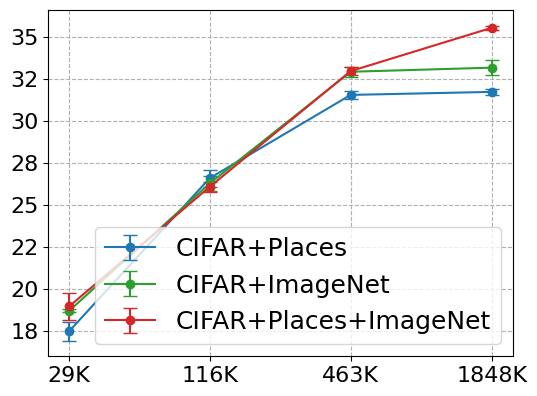}
    \end{subfigure} %
    \centering
    \begin{subfigure}{0.3\textwidth}
        \includegraphics[width=\linewidth]{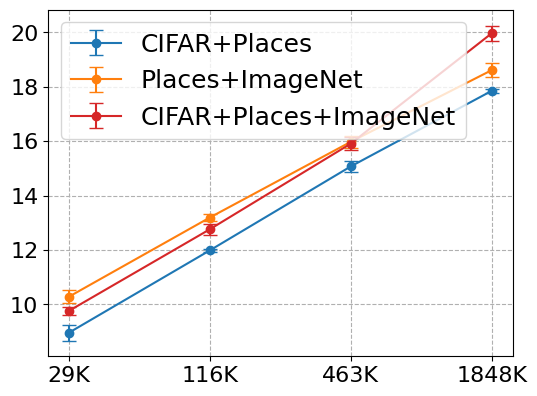}
    \end{subfigure} %
    \centering
    \begin{subfigure}{0.3\textwidth}
        \includegraphics[width=\linewidth]{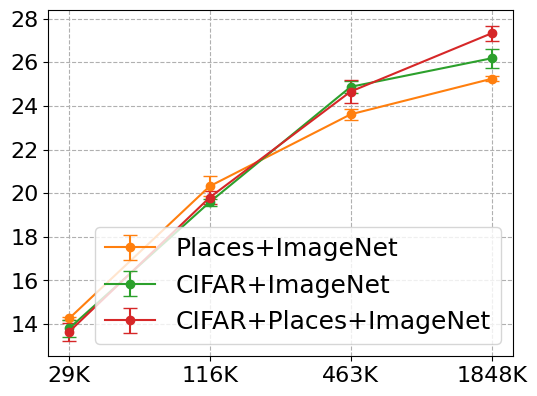}
    \end{subfigure} %
    
    \makebox[\textwidth]{\textbf{Number of Parameters (Log Scale)}}
    
    \caption{Our evaluation metrics tested with networks using Uncertainty loss weighting.}
    \label{fig:metrics-unc-sup}
\end{figure*}

\begin{figure*}[t]
    \centering
    \makebox[0.33\textwidth]{\textbf{CIFAR-100}}
    \makebox[0.3\textwidth]{\textbf{MiniPlaces}}
    \makebox[0.3\textwidth]{\textbf{Tiny-ImageNet}}
    \rotatebox[origin=lt]{90}{\makebox[0.1in]{\textbf{PerfGain [\%]}}}%
    \begin{subfigure}{0.3\textwidth}
        \includegraphics[width=\linewidth]{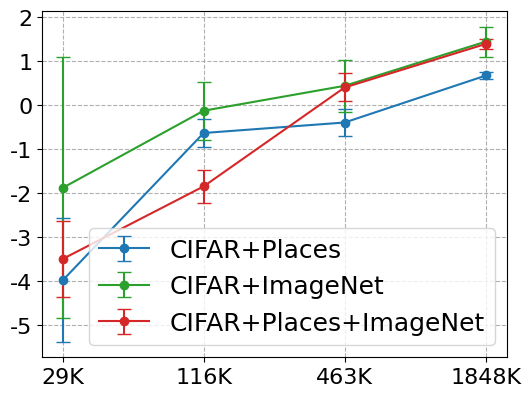}
    \end{subfigure} %
    \centering
    \begin{subfigure}{0.3\textwidth}
        \includegraphics[width=\linewidth]{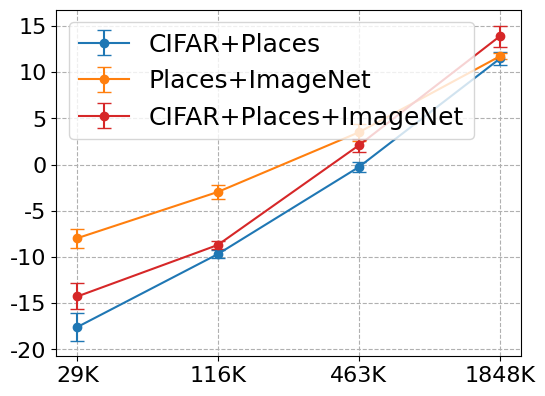}
    \end{subfigure} %
    \centering
    \begin{subfigure}{0.3\textwidth}
        \includegraphics[width=\linewidth]{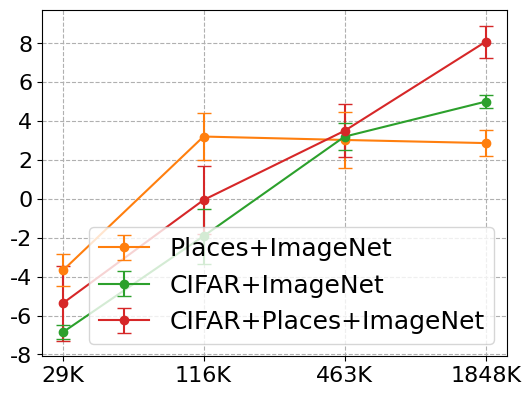}
    \end{subfigure} %

    \rotatebox[origin=lt]{90}{\makebox[0.1in]{\textbf{Interference [\%]}}}%
    \begin{subfigure}{0.3\textwidth}
        \includegraphics[width=\linewidth]{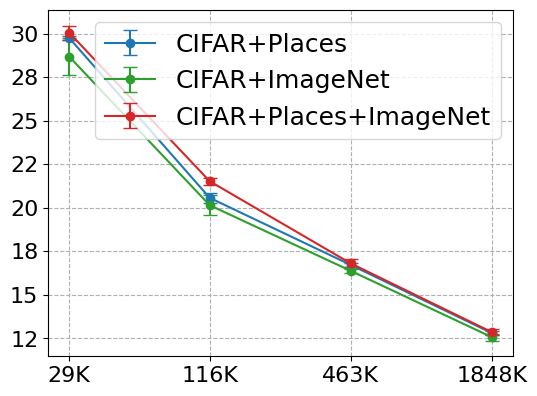}
    \end{subfigure} %
    \centering
    \begin{subfigure}{0.3\textwidth}
        \includegraphics[width=\linewidth]{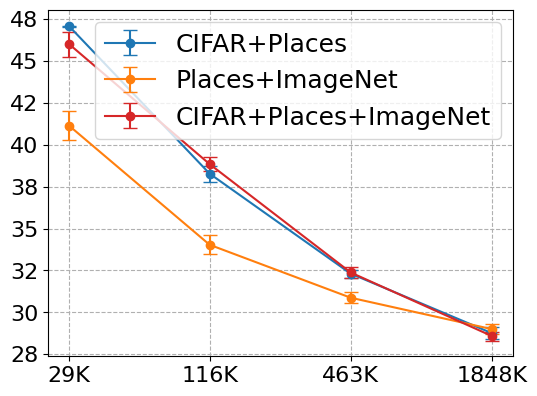}
    \end{subfigure} %
    \centering
    \begin{subfigure}{0.3\textwidth}
        \includegraphics[width=\linewidth]{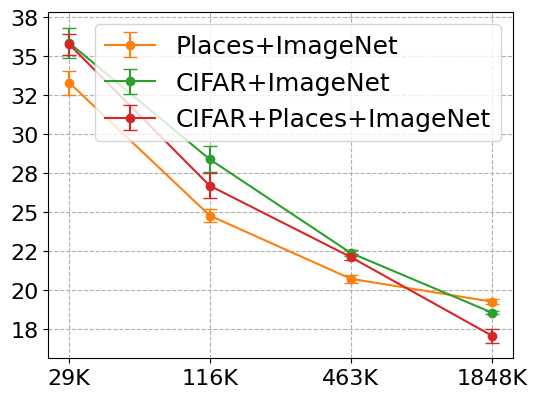}
    \end{subfigure} %
    
    \rotatebox[origin=lt]{90}{\makebox[0.1in]{\textbf{Transfer [\%]}}}%
    \begin{subfigure}{0.3\textwidth}
        \includegraphics[width=\linewidth]{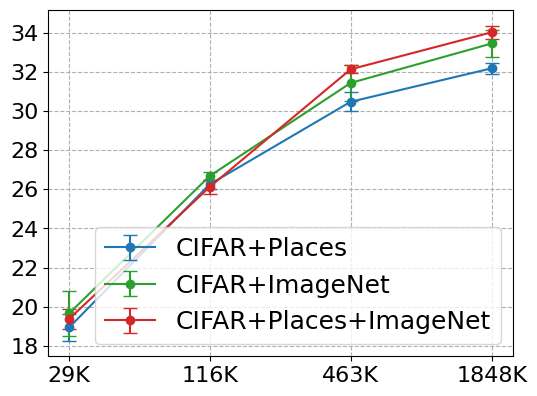}
    \end{subfigure} %
    \centering
    \begin{subfigure}{0.3\textwidth}
        \includegraphics[width=\linewidth]{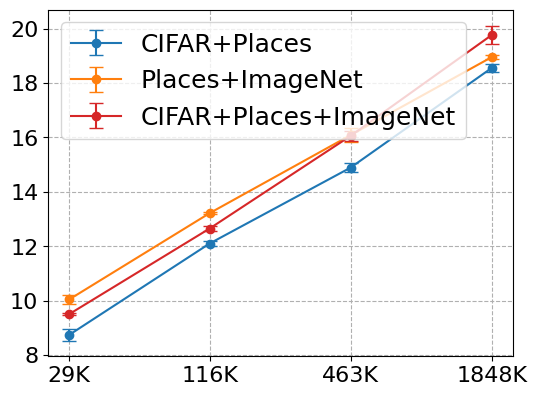}
    \end{subfigure} %
    \centering
    \begin{subfigure}{0.3\textwidth}
        \includegraphics[width=\linewidth]{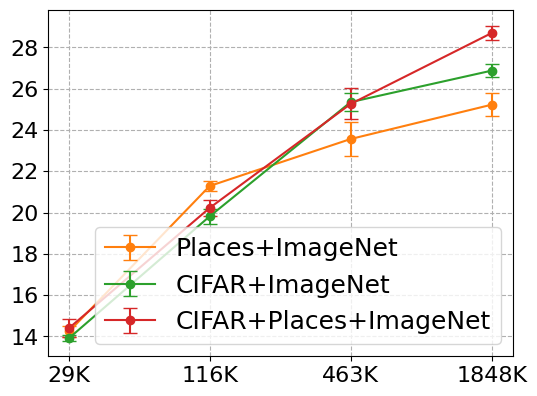}
    \end{subfigure} %
    
    \makebox[\textwidth]{\textbf{Number of Parameters (Log Scale)}}
    
    \caption{Our evaluation metrics tested with networks using CoV loss weighting. The last column is shown in the main text (Fig.~\ref{fig:metrics}).}
    \label{fig:metrics-cov-sup}
\end{figure*}

Our findings in the 2-domain setting still hold for the 3-domain models. However, 3-domain models exhibit more benefit from the largest capacity because of the need to learn representations general to all three domains. This is reflected by the fact that, as capacity grows, the 3-domain models' PerfGain increases, interference decreases, and transfer increases monotonically in all cases in Fig.~\ref{fig:metrics-naive-sup}, Fig.~\ref{fig:metrics-unc-sup}, and Fig.~\ref{fig:metrics-cov-sup}. Note that at the largest capacity, the 3-domain model always has better performance than the 2-domain model, except on CIFAR-100 using the Uniform weighting. However, we cannot make conclusions on the benefit of more domains because the 3-domain models are trained on more samples.

\end{document}